\documentclass{article}
\usepackage{newtxtext}
\usepackage[preprint]{neurips_2026}
\usepackage[utf8]{inputenc}
\usepackage[T1]{fontenc}
\usepackage{hyperref}
\usepackage{url}
\usepackage{booktabs}
\usepackage{amsfonts}
\usepackage{amsmath}
\usepackage{nicefrac}
\usepackage{microtype}
\usepackage{xcolor}
\usepackage{amssymb}
\usepackage{multirow}
\usepackage{xcolor,colortbl}
\usepackage{graphicx}
\usepackage{float}
\usepackage{algorithm}
\usepackage{algpseudocode}
\usepackage{adjustbox}
\usepackage{enumitem}
\usepackage{cleveref}
\usepackage{subcaption}
\usepackage{wrapfig}       
\usepackage{xspace}
\newcommand{\modelname}{\textsc{KnowsTFM}\xspace}

\title{KnowsTFM: Knowledge-Informed Fine-Tuning of \\
\emph{Small} Tabular Foundation Models}

\author{%
  Boshko Koloski\textsuperscript{1}
  \space Xiangjian Jiang\textsuperscript{2} 
  \space Senja Pollak\textsuperscript{1} \\ 
  \textbf{Bla\v{z} \v{S}krlj\textsuperscript{1} 
     \space Mateja Jamnik\textsuperscript{2} 
     \space Nikola Simidjievski\textsuperscript{3,2}
  } \\
  \textsuperscript{1} Jo\v{z}ef Stefan Institute and Postgraduate School, SI \\
  \textsuperscript{2}Department of Computer Science and Technology, University of Cambridge, UK \\
  \textsuperscript{3} Télécom Paris, Institut Polytechnique de Paris, FR \\
  \texttt{\{boshko.koloski, senja.pollak, blaz.skrlj@ijs.si\}} \\ 
  \texttt{\{xj265, ns779, mj201\}@cam.ac.uk} \\
  \texttt{nikola.simidjievski@telecom-paris.fr} 
}
\begin{document}
\maketitle

\begin{abstract}
%Tabular foundation models have pushed the standard for deep learning over tabular data by enabling strong default performance across many small and medium tabular tasks. However, in niche domains where data is scarce, high-dimensional, and shifted away from the pretraining distribution, they may still fail to outperform carefully designed domain-specific approaches. At the same time, domain experts have meticulously extracted relational knowledge into curated knowledge graphs and knowledge banks. This raises a central question: can such knowledge be used to improve and steer \textit{small} specialist tabular foundation models, especially when the target model must remain small, cheap to adapt, and practical to deploy? We study knowledge-informed fine-tuning of small tabular foundation models. Our focus is on nanoscale TabPFN- and TabICL-style variants that can be pretrained under controlled synthetic prior families and then adapted using two complementary mechanisms: structural attention priors derived from knowledge graphs and parameter-efficient low-rank updates. Our results show that injecting domain-specific structural knowledge during fine-tuning yields meaningful gains over vanilla variants in specialist settings, whereas in general-domain tasks the improvements are marginal. We further observe that continual fine-tuning of frontier models can trigger a collapse of their pretrained knowledge and mechanisms.
%%%%%%%%%%%%%%%%%%%%%%%%
%%%%% MATEJA's version
%%%%%%%%%%%%%%%%%%%%%%%%
Tabular foundation models have advanced deep learning for tabular data by delivering strong default performance across many small and medium tasks. Yet in niche domains, where data is scarce, high-dimensional, and shifted from the pretraining distribution, they may still fail to outperform carefully designed domain-specific methods. Many such domains also provide curated relational knowledge in the form of knowledge graphs and knowledge banks, but how to use this knowledge to improve and steer \textit{small} specialist tabular foundation models remains unclear.
We address this problem through \textbf{Know}ledge-informed fine-tuning of \textbf{s}mall \textbf{T}abular \textbf{F}oundation \textbf{M}odels (\modelname). Specifically, we study nanoscale TabPFN- and TabICL-style variants, pretrained under controlled synthetic prior families and adapted using two complementary mechanisms: structural attention priors derived from knowledge graphs and parameter-efficient low-rank updates. We show that injecting domain-specific structural knowledge during fine-tuning yields meaningful gains over vanilla variants in specialist settings, whereas gains on general-domain tasks are marginal. We further observe that continual fine-tuning of frontier models can trigger collapse of pretrained knowledge and mechanisms. 
\end{abstract}

\section{Introduction}

Tabular foundation models (TFMs) have emerged as a leading architecture for solving tabular machine learning tasks~\cite{qu2025tabicl,grinsztajn2025tabpfn,ma2026tabdpt}. These models are typically pretrained on large collections of synthetic tasks designed to mimic real-world datasets, enabling them to develop strong in-context learning capabilities. Instead of training a task-specific model from scratch, a pretrained TFM conditions on labelled support examples and predicts labels for query examples in a single forward pass. This paradigm is especially appealing for small-data regimes, where conventional deep learning models are difficult to train reliably.\looseness-1

At the same time, the broader foundation-model community has shown that small models can have an outsized impact. Accessible open weights models like Gemma~\cite{team2024gemma}, fully open-source initiatives such as OLMo~\cite{groeneveld-etal-2024-olmo}, and transparent, open-development Pythia-style~\cite{biderman2023pythia} models have democratised foundation-model research, making it reproducible and inspectable outside large industrial labs.  The recent nano TFM initiative, with nanoTabPFN~\cite{pfefferle2025nanotabpfn} and nanoTabICL~\cite{qu2025tabicl} as pioneering examples, has given similar opportunities to the tabular learning community, providing a sandbox environment for investigating the underlying mechanisms and exploring potential architectural optimisations.  We argue that {\em tabular} foundation models, and in particular small variants, require similar attention, specifically regarding how domain knowledge can be efficiently injected into these compact architectures
%they make repeated experimentation, domain adaptation, local deployment, and low-cost inference feasible.

Despite the promise of tabular foundation models, niche scientific domains remain difficult. Their pretrained priors are broad and generic, while domain data may be scarce, high-dimensional, and out-of-distribution. In biomedical prediction, for example, features may correspond to genes, drugs, pathways, proteins, or clinical entities, and relations among these entities are available in biomedical knowledge graphs. Such curated knowledge represents years of expert effort. Therefore, our central question is: \textit{Can formal knowledge structure be used to improve and steer small tabular foundation models without alleviating their pretrained in-context inference behaviour}.

\begin{figure}[!t]
    \centering
    \includegraphics[trim={0 2cm 0 0},clip,width=1\linewidth]{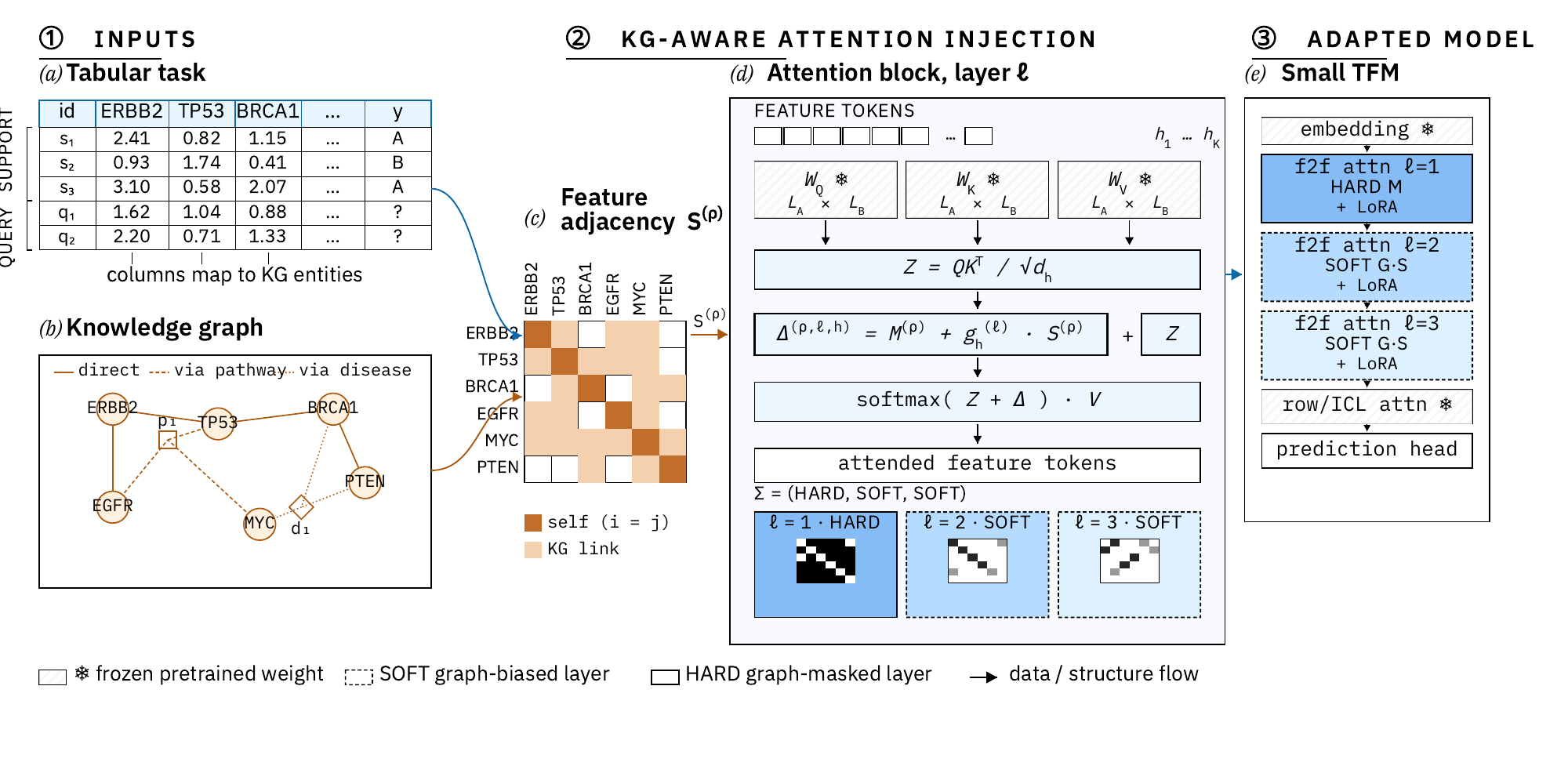}
    \caption{{\bf KG-aware fine-tuning of small tabular foundation models.} 
    Given tabular query features (a) and a knowledge graph (b), we derive a feature adjacency matrix from graph relationships (c), and inject this structure into transformer attention through graph-informed attention biases and hard/soft graph-masked layers (d). The resulting adapted small TFM combines frozen pretrained components with lightweight LoRA-based modules to produce downstream predictions (e).}
    \label{fig:kg-tab}
    \vspace{-5mm}
\end{figure}

We study this question through two knowledge-informed adaptation mechanisms, illustrated in \Cref{fig:kg-tab}, and evaluate them against stronger full-scale reference systems under practical cost constraints. First, we inject knowledge-graph structure into attention through hard masks or soft additive biases over feature-level interactions. Second, we use parameter-efficient low-rank adaptation to specialise small pretrained models while limiting destructive drift from the pretrained prior. This framework allows us to examine, in a controlled setting, how synthetic prior family, model scale, and KG-aware adaptation interact in small tabular foundation models. In this paper, we study knowledge-informed fine-tuning in the context of small tabular foundation models (TFMs), focusing on nanoscale TabPFN- and TabICL-style variants~\cite{pfefferle2025nanotabpfn}. These smaller models can be pretrained under controlled synthetic prior families, modified directly, and adapted at low cost. The overall aim is to make such models more useful in domain-specific settings while keeping them cheap enough to train, tune, benchmark, and deploy repeatedly.\looseness-1

%\paragraph{Contributions.}
Our main contributions are as follows:
\begin{enumerate}[topsep=-1pt, leftmargin=12pt, itemsep=0pt]
    \item 
    A small-model framework for tabular foundation modelling, with nanoscale TabPFN- and TabICL-style variants as controllable, low-cost targets for adaptation and full-scale models as ceiling/reference systems.
    \item 
    A knowledge-informed fine-tuning method for small TFMs that combines KG-derived structural attention priors with parameter-efficient low-rank adaptation.
    \item 
    Empirical evidence that KG-informed adaptation yields meaningful gains in specialist settings, limited gains in general-domain settings, and exposes the risk of pretrained knowledge collapse under continual fine-tuning.
\end{enumerate}

\section{Related Work}

Our work combines several research directions around a specific practical goal: tuning compact tabular foundation models to use external domain knowledge, reducing the performance gap to larger models while maintaining low adaptation and deployment costs. As such, it relates to prior work in tabular foundation models, domain-knowledge-informed prediction, structural priors, and parameter-efficient fine-tuning, which have been largely studied in isolation.

\textbf{Tabular foundation models.} Recent work has established foundation models as a viable paradigm for tabular deep learning~\cite{hollmann2022tabpfn,qu2025tabicl}. TabPFN showed that a transformer pretrained on synthetic tasks can solve small tabular classification problems in a single forward pass, effectively performing in-context learning over training examples~\citep{hollmann2022tabpfn}. Subsequent work broadened this design space with more general tabular transformers and transfer-oriented pretraining, including TransTab and CARTE~\citep{wang2022transtab,kim2024carte}, while TabPFN~v2 demonstrated that the foundation-model approach can scale to larger and more realistic settings~\citep{hollmann2025tabpfnv2}. TabICL extends in-context tabular modelling to larger datasets through a scalable TabICL-style architecture~\citep{qu2025tabicl}. On the adaptation side, TuneTables and LoCalPFN show that pretrained tabular models can be specialised efficiently~\citep{feuer2024tunetables,thomas2024localpfn}, and recent fine-tuning studies suggest that downstream adaptation improves performance but can materially alter the pretrained inference mechanism~\citep{rubachev2025finetuningtabular}. However, no adaptation work focused on using external knowledge to steer the fine-tuning for small TFMs. 

%\paragraph{Small foundation models and deployment.}
%The broader foundation-model community has demonstrated that small foundational models can strongly influence research and deployment practice~\cite{touvron2023llama}. Compact models make it possible to fine-tune locally, run repeated ablations, deploy on constrained hardware, and benchmark many variants at low cost~\cite{ding2025better}. This small-model paradigm motivates our work in the tabular setting. We ask whether small TFMs can play a similar role for structured tabular prediction: not replacing the largest models universally, but offering a cheap and adaptable operating point that can be specialised for domain-specific tasks, with the help of structured knowledge.
\textbf{Knowledge graphs for tabular and biomedical prediction.}
Knowledge-informed learning has also been studied in tabular and biomedical settings, although mostly through task-specific architectures rather than adaptation of pretrained tabular transformers. PLATO is close in spirit to our work: it uses an auxiliary knowledge graph to regularise high-dimensional tabular learning by tying feature weights through graph structure~\citep{ruiz2023plato} for the drug-to-gene response regression task. In biomedicine, DrugCell encodes Gene Ontology structure directly into a visible neural network for drug response prediction~\citep{kuenzi2020drugcell}, while DeepCDR and DRPreter combine graphs, pathway knowledge, and neural architectures for pharmacogenomic prediction~\citep{liu2020deepcdr,shin2022drpreter}. At the data level, resources such as Bioteque and PrimeKG further demonstrate the value of reusable biomedical knowledge graph representations for downstream prediction~\citep{duran2022bioteque,chandak2023primekg}. In contrast, our focus is not on designing a new task-specific architecture from scratch, but on injecting biomedical knowledge into a pretrained tabular foundation model through attention priors and parameter-efficient updates.\looseness-1

\textbf{Injecting structure into transformer attention.}
%A second line of work integrates external structure directly into transformer computation. 
In NLP, K-BERT introduces a visible matrix that restricts attention according to injected knowledge graph structure, and KnowBERT inserts knowledge modules between transformer layers to recontextualise token representations with entities~\citep{liu2020kbert,peters2019knowbert}. GREAT and GraphCodeBERT bias or mask attention using relational graphs over code~\citep{hellendoorn2020great,guo2021graphcodebert}. 
These works motivate our use of knowledge-graph-derived structural attention priors, but they do not address pretrained tabular foundation models, where features themselves may correspond to entities in a biomedical or semantic graph. %
LoRA and adapter-based transfer established that low-rank or bottleneck updates can adapt large pretrained transformers with only a small fraction of trainable parameters~\citep{hu2022lora,houlsby2019adapters}. Building on this idea, KnowLA and KG-Adapter integrate knowledge graph representations into LLM adaptation pipelines, while KBLaM modifies attention to incorporate external knowledge without fully updating the base model~\citep{luo2024knowla,tian2024kgadapter,wang2025kblam}. More deeply fused language--graph models such as GreaseLM and DRAGON further show that multi-layer interaction between transformer representations and graph reasoning can be beneficial~\citep{zhang2022greaselm,yasunaga2022dragon}. As summarised in the roadmap of~\cite{pan2024roadmap}, these methods form part of a broader trend towards knowledge-graph-enhanced foundation models. Our work extends this trend to the tabular domain.\looseness-1

%\textbf{Knowledge-steered parameter-efficient adaptation.}
%Our approach is also related to recent efforts to combine pretrained transformers, external knowledge, and parameter-efficient fine-tuning.

\textbf{Preserving in-context learning during fine-tuning.}
Finally, our use of parameter-efficient updates is motivated by recent evidence that naive fine-tuning can suppress in-context learning behaviour. ProMoT shows that conventional fine-tuning often induces format specialisation that harms generalisation~\citep{wang2024promot}, while~\cite{kotha2024implicitinference} argue that downstream tuning may not erase pretrained capabilities outright, but can bias the model away from invoking them. This issue is particularly relevant for TabPFN-style models, whose performance relies on pretrained in-context inference over tabular examples. Our method therefore uses lightweight updates and structured knowledge priors to adapt the model while reducing interference with its pretrained mechanism.

% \textbf{Positioning.} Taken together, prior work has studied tabular foundation models, knowledge-graph-informed biomedical predictors, structural attention priors, and parameter-efficient fine-tuning largely in isolation. Our work combines these directions around a specific practical goal: tuning small tabular foundation models so that they can use external domain knowledge and approach larger ceiling models while retaining low adaptation and deployment cost.

\section{Methodology}
\label{sec:method}
\vspace*{-2mm}
Let a tabular task consist of a labelled support set $\mathcal{D}_{s}=\{(x_i,y_i)\}_{i=1}^{n_s}$ and query inputs $\mathcal{D}_{q}=\{x_j\}_{j=1}^{n_q}$ (\Cref{fig:kg-tab}a). A pretrained tabular foundation model defines an in-context predictor $p_{\theta}(y_q \mid x_q, \mathcal{D}_s).$ We hypothesise that an external knowledge graph %$\mathcal{G}=(V,E)$ 
(e.g., PrimeKG~\cite{chandak2023primekg}, see \Cref{fig:kg-tab}b)  can provide an inductive prior over feature interactions (\Cref{fig:kg-tab}c), regularising adaptation in scarce-label regimes where the support set alone may not identify which columns should interact. We propose a three-component method: (\textbf{i}) controlled nanoscale TFMs, (\textbf{ii}) a KG-extraction pipeline that maps table columns to a feature-level interaction adjacency, (\textbf{iii}) \modelname a parameter-efficient KG-aware attention injection strategy (\Cref{fig:kg-tab}d-e). 
\subsection{Controlled small TFMs and synthetic priors}
%\textbf{Small TFMs and synthetic priors}
We pretrain TabPFN- and TabICL-style models as the intervention targets, following the nanoTabPFN recipe~\cite{pfefferle2025nanotabpfn} (see \Cref{app:pretrain_budget} for details). Each small model is defined by a backbone $b$, scale $s$, and synthetic prior family   $\theta = \theta(b,s,\pi)$.
Thus, two models with the same architecture but different pretraining priors are treated as distinct pretrained variants (see \Cref{tab:nanoscale_variants} in \Cref{app:pretrain_budget}). The synthetic prior family (see \Cref{tab:prior_families} in \Cref{app:pretrain_budget}) controls the distribution of tabular tasks observed during pretraining. We use two prior data generation regimes:\looseness-1
\[
    \pi_{\text{small}}: K \in [5,60],
    \qquad
    \pi_{\text{big}}: K \in [40,200],
\]
with a maximum sequence length of $300$ and at most $8$ classes in both cases.
The small and big prior regimes contain $1000$ and $2000$ generated batches, respectively, both drawn from the same Structural Causal Models (SCM)-based mixed prior family. Thus, they mainly differ in feature-count regime and the amount of generated prior data. By controlling the pretraining of small TFMs, we can test how differently sized synthetic priors generalise to truly unseen downstream tasks. All nano-scale variants in \Cref{tab:nanoscale_variants} use the same optimisation budget: $3000$ pretraining steps, batch size $4$, and a fixed random seed. The only pretraining input that changes across matched variants is $\pi \in \{\pi_{\text{small}},\pi_{\text{big}}\}$, so differences can be attributed to backbone family, scale, or synthetic prior regime rather than pretraining budget. This setup lets us examine whether KG-aware adaptation can compensate for weak or mismatched synthetic priors, make better use of support examples, or inject domain structure that scarce labels alone cannot reveal.\looseness-1

\subsection{Table to Knowledge Graph mapping} 
\label{sec:kg_mapping}
Let the tabular feature set be $\mathcal{C}=\{c_1,\dots,c_K\}$. We map columns to KG entities through a partial linking function $\phi:\mathcal{C}\rightarrow V\cup\{\varnothing\}$, where $\varnothing$ marks unmapped features. Based on the domain, we propose two mapping mechanisms:

\textbf{Grounding against expert-curated knowledge.} For biomedical datasets, gene-expression features can often be normalised to the HGNC-approved form (\texttt{HER2}~$\to$~\texttt{ERBB2}), and then linked to a gene-covering knowledge graph like PrimeKG~\citep{chandak2023primekg}.
Given the linked nodes, we define a biomedical feature-level adjacency by: 
% \[
% \resizebox{\textwidth}{!}{$\displaystyle
% S^{(\mathrm{bio})}_{ij}=1
% \Longleftrightarrow
% \underbrace{\{\phi(c_i),\phi(c_j)\}\in E_{\mathrm{gg}}}_{\text{direct gene--gene}}
% \lor
% \underbrace{\exists d\in V_{\mathrm{dis}}:
% \{\phi(c_i),d\}\in E_{\mathrm{gd}},\
% \{\phi(c_j),d\}\in E_{\mathrm{gd}}}_{\text{shared disease}}
% \lor
% \underbrace{\exists p\in V_{\mathrm{path}}:
% \{\phi(c_i),p\}\in E_{\mathrm{gp}},\
% \{\phi(c_j),p\}\in E_{\mathrm{gp}}}_{\text{shared pathway}}
% $}
% \]
\[
\begin{aligned}
\scriptstyle
S^{(\mathrm{bio})}_{ij}=1 \Longleftrightarrow\;&
\underbrace{\{\phi(c_i),\phi(c_j)\}\in E_{\mathrm{gg}}}_{\text{direct gene--gene}} \\
&\lor\underbrace{\exists d\in V_{\mathrm{dis}}:
\{\phi(c_i),d\}\in E_{\mathrm{gd}},
\{\phi(c_j),d\}\in E_{\mathrm{gd}}}_{\text{shared disease}} \\
&\lor\underbrace{\exists p\in V_{\mathrm{path}}:
\{\phi(c_i),p\}\in E_{\mathrm{gp}},
\{\phi(c_j),p\}\in E_{\mathrm{gp}}}_{\text{shared pathway}}
\end{aligned}
\]
where $E_{\text{gg}}$, $V_{\text{dis}}$, and $V_{\text{path}}$ denote PrimeKG
gene--gene edges, disease nodes, and pathway nodes, respectively. This setting
represents an idealised grounding regime: entity names are already close to
canonical identifiers, and the main challenge is graph projection rather than
entity disambiguation.

\textbf{General domain grounding.} General-domain tabular metadata is often messy or incomplete: curated KG with native typed entities may be unavailable, and column names are often abbreviated (\texttt{trestbps}, \texttt{Mg}, \texttt{chol}), making careful candidate generation and disambiguation essential. We therefore construct $S^{(\rho)}$ from Wikidata via a four-stage LLM-assisted mapping and disambiguation pipeline, executed once per dataset, that consumes only the dataset name and column descriptions:\looseness-1
\begin{enumerate}[topsep=-1pt, leftmargin=12pt, itemsep=0pt]
    \item \textbf{Query proposal.} A single Gemini-3 Pro Preview call sees the dataset name together with \emph{all} column descriptions simultaneously and $3$--$7$ Wikidata search queries per column. Cross-column conditioning is what makes this stage non-trivial: in a glass-composition dataset, \texttt{Mg} is proposed as ``magnesium'' rather than ``Madagascar'' precisely because the LLM
    sees the other element columns simultaneously.
    \item \textbf{Candidate retrieval.} Each proposed query is resolved against the Wikidata REST API; the top candidates are deduplicated and tagged with their unique identifiers QID. This step is deterministic and purely API-driven.
    \item \textbf{LLM disambiguation with abstention.} A second Gemini-3 Pro Preview call sees, per column, the original column description, the dataset context, and the candidate list, and either selects a single QID with a confidence and a rationale, or \emph{abstains}. The LLM is constrained to choose only from the retrieved candidates; it cannot invent a QID that is not in the Wikidata response. This abstention ensures no hallucination which might result from a naive ``prompt-the-LLM-for-the-QID'' baseline. %(Appendix~\ref{app:proflash} reports the Pro-vs-Flash
    %quality screen that justifies it).
    \item \textbf{Deterministic edge discovery.} Given the resolved QIDs, edges between feature pairs are obtained by SPARQL queries against the public Wikidata endpoint under three relation policies \mbox{(direct/1-hop/ancestor;} defined below). SPARQL is deterministic, so every step \emph{after} the LLM disambiguation is exactly reproducible from the cached \texttt{qids} artefact released with the paper.
\end{enumerate}

%\textbf{KG projection to feature space} For each pair of mapped columns $(c_{i},c_{j}$ we set A graph-construction policy $\rho$ maps the external KG to a feature-level adjacency $S^{(\rho)}\in\{0,1\}^{K\times K}$, where $S^{(\rho)}_{ij}=\mathbf{1}[R_{\rho}(\phi(c_i),\phi(c_j))]$ and $R_{\rho}$ is   a relation predicate determined by $\rho$.
%Across experimetns we consider the union between features. %For symmetric attention priors,  $S^{(\rho)} \leftarrow S^{(\rho)} \lor (S^{(\rho)})^\top \lor I_K$.  
%Density $\mathrm{dens}(S)=\|S\|_0/K^2$ is recorded for each dataset and used to construct density-matched random controls $\widetilde{S}$.

%Mapping quality is summarised by coverage $\mathrm{cov}(\phi)= K^{-1}\sum_{i=1}^{K}\mathbf{1}[\phi(c_i)\neq\varnothing]$. For biomedical datasets we use direct gene-symbol matching to PrimeKG~\citep{chandak2023primekg} via PyKEEN. For general-domain and
%Bayesian-network datasets we use a three-stage agentic pipeline: an LLM proposes 3--7 search queries per column conditioned on the dataset description, deterministic Wikidata search returns top-$K$ candidates with type information, and a second LLM call selects the best QID per column or abstains. The pipeline is uniform across domains; only the underlying KG changes.

\subsection{\modelname: KG-aware parameter-efficient injection}
\label{sec:method_injection}
The resulting structural prior $S^{(\rho)}$ is injected into the pretrained backbone through a low-rank parameter-efficient attention modification. The main
idea is to add a graph-derived term to the pre-softmax logits of feature-attention layers, so that attention is biased toward feature pairs connected in the knowledge graph. This gives a KG-aware LoRA attention fine-tuning strategy: the pretrained backbone and the KG adjacency remain frozen, while only LoRA factors and a small number of KG-bias parameters are learned. The full algorithm breakdown can be found in \Cref{app:kglora}.

For a feature-attention layer with queries $Q$, keys $K$, and values $V$, the
standard attention logits are:
\[
    Z=\frac{QK^\top}{\sqrt{d_h}}.
\]
We inject the projected graph prior as an additive logit modification before the
softmax:
\[
    \mathrm{Attn}_{\mathrm{KG}}(Q,K,V)
    =
    \operatorname{softmax}
    \left(
        Z + \Delta^{(\rho)}
    \right)V.
\]
The form of $\Delta^{(\rho)}$ depends on the layer slot, which we define next.

\textbf{Per-layer schedule and per-head soft bias.} Rather than using a single global injection strength, we assign to each feature-attention block $\ell$ a slot $ \sigma_\ell \in \{\textsc{hard},\textsc{soft},\textsc{off}\}$, thus yielding $\boldsymbol{\sigma}=(\sigma_1,\dots,\sigma_L).$ A \textsc{hard} slot applies a graph mask:
\[
    M^{(\rho)}_{ij}
    =
    \begin{cases}
    0, & S^{(\rho)}_{ij}=1,\\
    -C, & S^{(\rho)}_{ij}=0,
    \end{cases}
    \qquad C\gg 0,
\]
so that off-graph feature pairs are suppressed in the attention logits. A
\textsc{soft} slot adds a learnable per-head graph bias:
\[
    \Delta^{(\rho,\ell,h)}_{\textsc{soft}}
    =
    g_h^{(\ell)} S^{(\rho)}, \quad\quad \text{where} \quad\quad
    g_h^{(\ell)}
    =
    \sigma(a_h^{(\ell)})\,s_h^{(\ell)}.
\]
Here $a_h^{(\ell)}$ is a learned gate parameter and $s_h^{(\ell)}$ is a learned
scale parameter. They are initialised as $a_h^{(\ell)}=0$ and
$s_h^{(\ell)}=1$, so the initial soft bias strength is
$g_h^{(\ell)}=0.5$. An \textsc{off} slot leaves the attention logits unchanged. Our main schedule is  
$\boldsymbol{\sigma} =(\textsc{hard},\textsc{soft},\textsc{soft})$, extended as needed for deeper backbones. Thus, the first adapted block enforces
the KG mask, while later blocks use softer KG-guided biases. The KG enters only
as an additive term to the pre-softmax attention logits; it is not injected into
the value vectors, embedding layer, or feed-forward layers.

\textbf{LoRA-augmented projections.}
KG priors constrain attention, but task adaptation still requires trainable
capacity. For a pretrained projection matrix
$W\in\mathbb{R}^{d_{\mathrm{out}}\times d_{\mathrm{in}}}$, LoRA defines:
\[
    W_{\mathrm{eff}}
    =
    W+\frac{\alpha}{r}L_B L_A,
    \qquad
    L_A\in\mathbb{R}^{r\times d_{\mathrm{in}}},
    \quad
    L_B\in\mathbb{R}^{d_{\mathrm{out}}\times r}.
\]
The base weight $W$ is frozen, and only $L_A$ and $L_B$ are learned. We attach
LoRA to every linear projection of each adapted block: the joint QKV projection,
the attention output projection, and both linear layers of the feed-forward MLP.
In all experiments, we use rank $r=16$ and scaling $\alpha=32$. During fine-tuning, the trainable parameters are conditioned under LoRA updates and the soft-slot gate and scale parameters. The pretrained backbone weights,
the graph adjacency $S^{(\rho)}$, and the hard mask derived from it remain frozen. This separates external structure from trainable capacity: the KG determines which feature interactions are encouraged or suppressed, while LoRA provides the task-specific adaptation capacity.
%The ablation arms then test, at fixed schedule and fixed graph source, whether LoRA capacity alone (LoRA FT, no KG) suffices and whether combining structure with low-rank tuning (KG-LoRa FT) is complementary.

%\textbf{Backbone-specific injection} The injection site is architecture-dependent. If the model exposes feature tokens or column-attention blocks, $S^{(\rho)}$, $B^{(\rho)}$, and $M^{(\rho)}$ are applied directly to feature attention. If the model primarily attends over rows or support-query examples, KG structure is injected through column representations or through LoRA-modified attention projections. This is
%why small TFMs are useful: their adaptation surface is inspectable, while full TabPFN v2.6 and TabICL v2 remain ceiling models.

\section{Experimental Setup}
\label{sec:experimental_setup}
The experimental analysis is organised around three questions, each addressing a different aspect in when external knowledge improves adaptation of tabular foundational models, and in particular, small variants: domain adaptability  (Q1), generalisability (Q2) and structure recovery (Q3).

\textbf{Q1: Does knowledge-informed tuning improve TFMs on a specific domain?} For the main experimental track, we focus on the gene-expression classification of the CUMIDA-42~\cite{cumida2019} suite of $42$ gene microarray problems. In this setting, feature columns correspond to gene symbols and can be directly matched to PrimeKG (\Cref{sec:kg_mapping}). For each of the eight nanoscale models (see~\Cref{tab:nanoscale_variants} in \Cref{app:pretrain_budget}), as well as for the two frontier baselines TabPFN~v2.6 and TabICL~v2, we compare KG-aware fine-tuning (\modelname FT) against vanilla fine-tuning (Vanilla FT), with pretrained zero-shot inference as a baseline. For this benchmark, we design experimental arms consisting of 100 features each. We vary the proportion of informative features to uninformative features (see \Cref{app:dataset_meta}), specifically testing intervals of 0\% (all random), 25\% (noisy), 50\% (headline), and 100\%  (clean). Our headline results focus on a balanced 50-50 split. The primary motivation is to evaluate the importance of introducing domain-specific knowledge. We hypothesise that external structured knowledge is unnecessary for a fully curated set of features (100\% split), whereas it proves highly beneficial when dealing with completely random features (0\% case).  

\textbf{Q2: Does knowledge-informed tuning make TFMs competitive in general?} We are interested in whether our proposed tuning method helps bridge the gap to the zero-shot frontier models. To evaluate that, we construct a set of $13$ datasets originating from different domains (see \Cref{app:dataset_meta}). We use the LLM-assisted grounding described in \Cref{sec:kg_mapping}, to derive the feature graph from Wikidata. As a knowledge-structure control, we also consider an external LLM (Gemini-3.0 Pro Preview), acting as an oracle, to produce a feature graph between the datasets (see \Cref{app:llm_prompts}). This tests whether the benefit comes from grounded external structure or from the LLM's implicit knowledge alone. Additionally, on 13 datasets, we test whether knowledge-informed tuning can improve the performance of small TFMs, under default and fine-tuned settings. \looseness-1 %We compare the Wikidata's (KG) and Gemini's (LLM) graph in the default KnowsTFM. \looseness-1

%LLM-derived graph constructed without Wikidata grounding; comparing it against extracted KG injection at base recipe setting. This tests whether the benefit comes from grounded external structure or from the LLM's implicit knowledge alone.\looseness-1

\textbf{Q3: How well can we recover structure by querying a general knowledge graph?} 
For three datasets from the TabStruct \citep{jiang2026tabstruct} suite with available ground-truth graphs, we fix the schedule and LoRA hyperparameters and vary only the graph prior $S^{(\rho)}$. We compare: (i)~the ground-truth \textbf{DAG} in the Bayesian-network setting; (ii)~\textbf{Wikidata}-derived graphs under the direct, 1-hop, ancestor, and union relation policies; and (iii)~a density-matched \textbf{random} graph $\widetilde{S}$. This separates meaningful graph semantics from the effect of sparse attention regularisation alone.

% \textbf{Q4: How does model scale and prior size interact with KG-aware adaptation?}
% We compare nanoscale TabPFN- and TabICL-style models against TabPFN v2.6 and TabICL v2. This tests whether external graph structure is most useful for small models with limited synthetic-prior coverage, or whether it also improves larger frontier tabular foundation models.

%\textbf{\color{blue}ADD DISCUSSION ON MODEL VARIANTS The differences in priors and architecture i.e what "big/small" and "small/base" means. Refer the reader to the table in the appendix E (?) with the configurations}

%\subsection{Protocol and reporting}
%\label{sec:setup:protocol}
\textbf{Protocol and reporting.} For each experiment, we use $3$ random seeds and $3$ stratified folds, with feature scaling fitted separately within each training fold. We report balanced accuracy as the primary metric. Both Vanilla FT and \modelname (denoting KG-LoRA tuning) are fine-tuned with early stopping on a fold-local validation set comprising $20\%$ of the training split. We use the same optimisation budget for both methods (at most $500$ steps, patience $25$, evaluation every $10$ steps, AdamW~\cite{loshchilov2018decoupled}, and weight decay $10^{-5}$). For the biomedical experiments, we use default settings without hyperparameter tuning. For Vanilla FT, we set $\mathrm{lr}=10^{-4}$. For \modelname, we set $\mathrm{lr}=10^{-3}$ and use a single rollout schedule $\boldsymbol{\sigma}=(\textsc{hard},\textsc{soft},\textsc{soft})$. For the general-domain experiments (Q2), we tune each method using its method-specific hyperparameters. More implementation details are available in \Cref{sec:implementation}.

\section{Results}
\textbf{Knowledge-informed tuning improves performance on domain-specific tasks (Q1)}. We report the primary results on CUMIDA-42 in \Cref{tab:cumida42_no_rnd}. First, for small model variants, our knowledge-informed fine-tuning strategy can lead to substantial downstream performance improvements over their zero-shot performance, in some cases up to 30\%. The performance improvements achieved with \modelname are also evident when compared to a standard fine-tuning (Vanilla FT) strategy. While less substantial ($\sim$ 3\%), we find that these improvements are consistent across all eight nanoscale variants throughout the benchmarked datasets. Compared to LoRA fine-tuning, \modelname also generally leads to improvements, albeit with a smaller gap. The improvement over LoRA-only FT further isolates the structural contribution of the KG: low-rank adaptation without the PrimeKG-derived attention prior explains less than half of the total gain. This is exactly the setting where the proposed mechanism should excel: CUMIDA columns are gene symbols, the feature-level graph is derived from a curated and biologically meaningful KG, and the support sets are small relative to the feature dimension, making it difficult to infer reliable feature interactions from labels alone. All of these results support our main thesis, that for domain-specific tasks, knowledge-informed fine-tuning is beneficial. \looseness-1

\begin{table}[t]
\centering\footnotesize
\caption{Mean balanced accuracy on the CUMIDA-42 gene-expression benchmark. \textbf{Bold} = best in row. \emph{Pretrained}: zero-shot in-context inference. \emph{Vanilla FT}: full-model episodic fine-tuning. \emph{LoRA FT}: tuning without graph mask.  \modelname: our proposed method.}
\label{tab:cumida42_no_rnd}
\resizebox{0.9\textwidth}{!}{
    \begin{tabular}{@{}lrcccc@{}}
    \toprule
    Model & Params & Zero-shot & Vanilla FT & LoRA FT & \modelname  \\
    \midrule
    NanoTabICL big-prior, base & 3.30 M & 0.582$_{\pm 0.239}$ & 0.772$_{\pm 0.145}$ & \textbf{0.796$_{\pm 0.141}$} & 0.793$_{\pm 0.139}$ \\
    NanoTabICL big-prior, small & 0.85 M & 0.543$_{\pm 0.210}$ & 0.748$_{\pm 0.141}$ & 0.758$_{\pm 0.137}$ & \textbf{0.767$_{\pm 0.145}$} \\
    NanoTabICL small-prior, base & 3.30 M & 0.649$_{\pm 0.257}$ & 0.792$_{\pm 0.140}$ & 0.808$_{\pm 0.142}$ & \textbf{0.815$_{\pm 0.137}$} \\
    NanoTabICL small-prior, small & 0.85 M & 0.475$_{\pm 0.145}$ & 0.751$_{\pm 0.148}$ & 0.767$_{\pm 0.140}$ & \textbf{0.768$_{\pm 0.151}$} \\
    NanoTabPFN big-prior, base & 3.72 M & 0.427$_{\pm 0.118}$ & 0.733$_{\pm 0.138}$ & 0.723$_{\pm 0.140}$ & \textbf{0.747$_{\pm 0.135}$} \\
    NanoTabPFN big-prior, small & 1.13 M & 0.441$_{\pm 0.117}$ & 0.706$_{\pm 0.137}$ & 0.719$_{\pm 0.133}$ & \textbf{0.757$_{\pm 0.150}$} \\
    NanoTabPFN small-prior, base & 3.72 M & 0.443$_{\pm 0.107}$ & 0.724$_{\pm 0.145}$ & 0.720$_{\pm 0.139}$ & \textbf{0.733$_{\pm 0.148}$} \\
    NanoTabPFN small-prior, small & 1.13 M & 0.448$_{\pm 0.110}$ & 0.704$_{\pm 0.144}$ & 0.720$_{\pm 0.145}$ & \textbf{0.767$_{\pm 0.139}$} \\
    \midrule
    \rowcolor{gray!12} \textit{mean over 8 nano ckpts} & --- & 0.501$_{\pm 0.187}$ & 0.741$_{\pm 0.144}$ & 0.751$_{\pm 0.142}$ & \textbf{0.768$_{\pm 0.144}$} \\
    \rowcolor{gray!12} \textit{$\Delta$ vs Vanilla FT} & --- & -0.240 & --- & +0.010 & \textbf{+0.027} \\
    \midrule
    \rowcolor{blue!8} TabICL v2 & 27.05 M & \textbf{0.893$_{\pm 0.114}$} & 0.884$_{\pm 0.116}$ & 0.884$_{\pm 0.116}$ & 0.876$_{\pm 0.120}$ \\
    \rowcolor{blue!8} TabPFN v2.6 & 10.73 M & \textbf{0.892$_{\pm 0.118}$} & 0.862$_{\pm 0.127}$ & 0.857$_{\pm 0.128}$ & 0.865$_{\pm 0.121}$ \\
    \bottomrule
    \end{tabular}
}
\vspace{-5mm}
\end{table}

To further evaluate the effect of the knowledge injection, we investigate the performance across different feature-noise setups \Cref{tab:cumida_arms_main_summary}. Across all four setups, our \modelname achieves the best mean and rank score. This points to the robustness of our method despite the input-level noise. While absolute performance generally degrades as noise increases, we don't observe a similar trend in the relative performance gain over vanilla fine-tuning. This further highlights the potential of knowledge-aware attention steering and raises the question of whether greater gains could be achieved by more sophisticated relation aggregation prior to adjacency building, rather than the current rule-based aggregation (see \Cref{app:attention_viz}). A detailed breakdown of results and experiments is available in \Cref{sec:feat_arm}. \looseness -1

\begin{figure}[b]
\vspace{-7mm}
    \centering
    \includegraphics[width=0.88\linewidth]{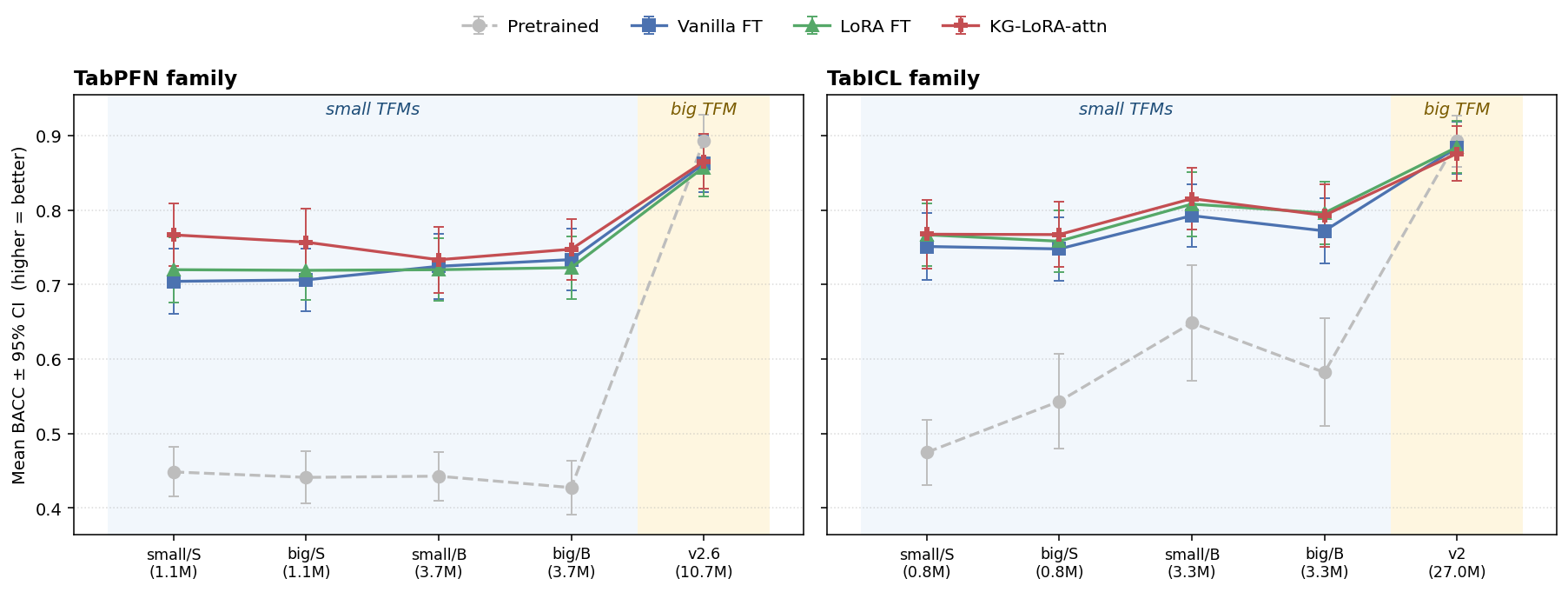}
    \caption{Side-by-side comparison of performance across different models and priors. Priors are divided into small and large groups, where \textit{B} is the base model and \textit{S} is the small model. Small tabular models are shown in the blue subplot, while the light orange subplot denotes the large models.}
\label{fig:model_comparison}
\end{figure}

Across backbones, the NanoTabPFN model benefits more from the KG term than NanoTabICL (\Cref{fig:model_comparison}). The average KG over LoRA-only gain is nearly double for NanoTabPFN compared to NanoTabICL, implying that prior contributes signal mainly in the PFN-style backbone. We attribute this to the simpler feature interactions in the PFN model, whereas NanoTabICL's shared row/column computation already provides a more structured adaptation structure. Smaller models also benefit more from the KG injection, relative to the base models. We interpret this as a capacity effect: under the current training regime, smaller models have less ability to infer useful feature interactions than larger models, making the external KG prior more beneficial. This pattern may change under longer training runs. Across structural prior sizes, we do not observe any meaningful differences, with the larger prior yielding a larger relative improvement.

% Our \modelname improves the performance of small-TFM compared to standard fine-tuning  (Vanilla FT). We find that these improvements are consistent across all eight nanoscale variants, 

% (\Cref{fig:model_comparison}). 

% checkpoints; indeed, in seven of the eight checkpoints, our method is the best-performing small-TFM method. The improvement over LoRA-only FT further isolates the structural contribution of the KG: low-rank adaptation without the PrimeKG-derived attention prior explains less than half of the total gain. This is exactly the setting where the proposed mechanism should excel: CUMIDA columns are gene symbols, the feature-level graph is derived from a curated and biologically meaningful KG, and the support sets are small relative to the feature dimension, making it difficult to infer reliable feature interactions from labels alone. 

Surprisingly, both frontier models degrade when adapted to these small datasets, as shown in \Cref{tab:cumida42_no_rnd} and the orange subplots of \Cref{fig:model_comparison}. This trend is consistent regardless of the tuning method. We interpret this as a possible scale-and-data mismatch effect: {TabPFN v2.6} and {TabICL v2} already encode broad tabular structure through large-scale synthetic-prior pretraining. Fine-tuning them on a single small target dataset can disturb this robust in-context behaviour rather than improve it. This resembles phenomena observed in large language models, where fine-tuning on narrow or newly introduced knowledge can degrade the use of pretrained knowledge and increase the tendency toward hallucination~\citep{gekhman2024does, li-etal-2024-revisiting}. Mitigating this degradation may require mixed-data fine-tuning, stronger regularisation, more careful hyperparameter optimisation and more thoughtful fine-tuning setup implementation~\cite{rubachev2025finetuningtabular}.\looseness-1

%\begin{wraptable}{r}{0.58\textwidth}

\textbf{General knowledge graphs can lead to improvements and bridge the scale gap (Q2).} Next, we investigate whether the findings on the domain-specific tasks can also translate to more general tasks using general-purpose knowledge graphs. Our results (\Cref{tab:xdomain13_adaptation_regimes}) show that when using a Wikidata-grounded feature graph, the performance of small TFMs using \modelname, while still substantially better than the zero-shot variants, is comparable to that of standard vanilla fine-tuning. This suggests that grounding using a general-purpose KGs might not provide sufficient gain, beyond the one from the data. We further evaluate knowledge-grounding by directly comparing with an LLM variant $\textsc{LLM-LoRA-attn~FT}$ (\Cref{tab:cross_domain_13ds}), which uses the same adaptation pipeline but replaces Wikidata-grounded edges with the LLM's own proposed feature--feature graph. We find that, in this case, the performance of small TFM regresses compared to both \modelname and vanilla strategies. We attribute this to the lower density of the derived graphs, highlighting the importance of formal knowledge graph grounding. \looseness -1

\begin{table}[t]
\vspace{-5mm}
    \begin{minipage}[t]{0.48\linewidth}
        \caption{Mean performance of our approach compared to vanilla fine-tuning on the CUMIDA42 dataset, across the four noise level feature attributions. The knowledge-steered tuning allows for improved performance across all noise levels. $^{***} $ indicates statistical significance. \looseness=-1}
        \label{tab:cumida_arms_main_summary}
        \begin{adjustbox}{max width=\linewidth}
{\renewcommand{\arraystretch}{1.22}
\setlength{\extrarowheight}{1.5pt}
\begin{tabular}{@{}lccc@{}}
\toprule
Dataset & Zero-shot & vFT & \modelname  \\
\midrule
\multirow{2}{*}{\texttt{top100} } & $0.505\,{\scriptstyle\pm 0.150}$ & $0.746\,{\scriptstyle\pm 0.136}$ & $\mathbf{0.789}\,{\scriptstyle\pm 0.138}$~${}^{***}$ \\
(clean) & rank $2.95$ & rank $1.93$ & rank $\mathbf{1.12}$ \\
\midrule
\multirow{2}{*}{\texttt{top50\_rnd50} } & $0.501\,{\scriptstyle\pm 0.148}$ & $0.742\,{\scriptstyle\pm 0.137}$ & $\mathbf{0.768}\,{\scriptstyle\pm 0.140}$~${}^{***}$ \\
(headline) & rank $2.95$ & rank $1.93$ & rank $\mathbf{1.12}$ \\
\midrule
\multirow{2}{*}{\texttt{top25\_rnd75} } & $0.489\,{\scriptstyle\pm 0.141}$ & $0.711\,{\scriptstyle\pm 0.144}$ & $\mathbf{0.747}\,{\scriptstyle\pm 0.142}$~${}^{***}$ \\
(noisy) & rank $2.90$ & rank $1.99$ & rank $\mathbf{1.11}$ \\
\midrule
\multirow{2}{*}{\texttt{rnd100}} & $0.479\,{\scriptstyle\pm 0.135}$ & $0.633\,{\scriptstyle\pm 0.150}$ & $\mathbf{0.684}\,{\scriptstyle\pm 0.153}$~${}^{***}$ \\
(all-rnd) & rank $2.86$ & rank $1.93$ & rank $\mathbf{1.21}$ \\
\bottomrule
\end{tabular}}
\end{adjustbox}
    \end{minipage}
    \hfill
    \begin{minipage}[t]{0.49\linewidth}
        \caption{Results for NanoTabICL and NanoTabPFN (both big-prior, base), across adaptation regimes. Further hyperparameter optimisation narrows the gap between the frontier model and ours on general domain datasets.}
        \label{tab:xdomain13_adaptation_regimes}
        \resizebox{\linewidth}{!}{
        \begin{tabular}{@{}lr@{}}
        \toprule
        Adaptation regime & xdomain13 BACC \\
        \midrule
        \multicolumn{2}{@{}l}{\textit{NanoTabICL \;\; (3.30\,M)}} \\
        Pretrained (zero-shot) & $0.413 {\scriptstyle\pm 0.144}$ \\
        Vanilla FT (default recipe) & $0.728 {\scriptstyle\pm 0.135}$ \\
        Vanilla FT (HPO) & $0.722 {\scriptstyle\pm 0.124}$ \\
        \modelname  FT (default recipe) & $0.726 {\scriptstyle\pm 0.132}$ \\
        \textbf{\modelname (HPO)} & $\mathbf{0.730} {\scriptstyle\pm 0.125}$ \\
        \midrule
        \multicolumn{2}{@{}l}{\textit{NanoTabPFN, base \;\; (3.72\,M)}} \\
        Pretrained (zero-shot) & $0.408 {\scriptstyle\pm 0.151}$ \\
        Vanilla FT (default recipe) & $0.708 {\scriptstyle\pm 0.129}$ \\
        Vanilla FT (HPO) & $0.707 {\scriptstyle\pm 0.135}$ \\
        \modelname  (default recipe) & $0.697 {\scriptstyle\pm 0.136}$ \\
        \textbf{\modelname (HPO)} & $\mathbf{0.724} {\scriptstyle\pm 0.113}$ \\
        \midrule
        \rowcolor{blue!8} TabICL v2 \;\; (27.05\,M) Pretrained (zero-shot) & $\mathit{0.763} {\scriptstyle\pm 0.141}$ \\
        \rowcolor{blue!8} TabPFN v2.6 \;\; (10.73\,M) Pretrained (zero-shot) & $\mathit{0.764} {\scriptstyle\pm 0.141}$ \\
        \bottomrule
        \end{tabular}}
    \end{minipage}
    \vspace{-5mm}
\end{table}

\begin{table}[b]
\centering\footnotesize
\vspace{-7mm}
\caption{Mean balanced accuracy on \textbf{13 non-biomedical datasets} with agentically-mapped Wikidata KGs, at $3$ seeds $\times$ $3$ folds. \emph{\modelname} uses the Wikidata-grounded $S^{(\rho)}$; \emph{LLM-LoRA-attn FT} uses Gemini-proposed feature--feature edges with \emph{no} Wikidata grounding. We use the default schedule  $\boldsymbol{\sigma}$.}
%\textbf{Bold} = best small-TFM method per row.\looseness-1}
\label{tab:cross_domain_13ds}
\resizebox{!}{!}{
\begin{tabular}{@{}lcccc@{}}
\toprule
Model &  Zero-shot & Vanilla FT & \modelname & LLM-LoRA-attn FT \\
\midrule
NanoTabICL big-prior, base & $0.414_{\pm 0.144}$ & {$\mathbf{0.728_{\pm 0.135}}$} & $0.726_{\pm 0.132}$ & $0.616_{\pm 0.188}$ \\
NanoTabPFN big-prior, base &  $0.408_{\pm 0.151}$ & {$\mathbf{0.708_{\pm 0.129}}$} & $0.697_{\pm 0.136}$ & $0.435_{\pm 0.171}$ \\
\midrule
% FIXED: Removed the extra "--- &" so this row has exactly 5 columns
\rowcolor{gray!12} \textit{mean over 2 ckpts} & $0.411_{\pm 0.147}$ & {$\mathbf{0.718_{\pm 0.131}}$} & $0.712_{\pm 0.131}$ & $0.525_{\pm 0.153}$ \\
% FIXED: Removed the extra "--- &" so this row has exactly 5 columns
\rowcolor{gray!12} \textit{$\Delta$ vs Vanilla FT} & $-0.308$ & --- & $-0.007$ & $-0.193$ \\
\midrule
\rowcolor{blue!8} TabICL v2 & \textbf{$\mathbf{0.763_{\pm 0.141}}$} & $0.743_{\pm 0.129}$ & $0.749_{\pm 0.131}$ & $0.722_{\pm 0.137}$ \\
\rowcolor{blue!8} TabPFN v2.6 &  $\mathbf{0.764_{\pm 0.141}}$ & $0.750_{\pm 0.129}$ & $0.750_{\pm 0.123}$ & $0.408_{\pm 0.150}$ \\
\bottomrule
\end{tabular}
}
\end{table}

Finally, similarly to the domain-specific tasks, we find that the big models again lose accuracy under fine-tuning (\Cref{tab:xdomain13_adaptation_regimes} and \Cref{sec:appendix:xdomain13_per_dataset}): TabICL~v2 drops from $0.763$ zero-shot to $0.743$ under Vanilla FT and $0.749$ under \modelname, while TabPFN~v2.6 drops from $0.764$ zero-shot to $0.750$ under both Vanilla FT and \modelname.  On the other hand, we find that when hyperparameters are tuned, for instance, with different strategies, we can further improve the performance of \modelname variants (\Cref{tab:xdomain13_kg_winners} in \Cref{sec:xdomain13_kg_winners}). More importantly, HPO further reduces the gap between the \modelname small TFM variants and their large counterparts.  The larger frontier models still outperform them, but only by a margin of 4\%. We conjecture that failure to improve performance further on some datasets can be attributed to uneven coverage of mapped columns, mapping ambiguity that propagates into the built adjacency, and low domain coverage of the graph (see \Cref{tab:kg_llm_stats} and  \Cref{app:kg-stats}).\looseness-1

%\Cref{} in
% fine-tuned, we can further improve the average performance of \modelname. For the base model in the NanoTabPFN family, using the default KG-tuning recipe actually degrades performance compared to Vanilla Fine-Tuning (FT). However, once HPO is performed, performance improves beyond that of the vanilla FT HPO variant. Despite the gains achieved from fine-tuning the 3M-parameter models, the larger frontier models still outperform them by a margin of 4\% BACC.  Failure to improve performance on some datasets can be attributed to uneven coverage of mapped columns, mapping ambiguity that propagates into the built adjacency, and low domain coverage of the graph (for details see \Cref{tab:kg_llm_stats} in \Cref{app:kg-stats}).\looseness-1

% In the default architectural setting, on the noisier and more realistic general domain datasets, 

% the results, expectedly, drift from the biomedical setting (\Cref{tab:cross_domain_13ds}). Under the same schedule, the small-TFMs' balanced accuracy is almost tied with vanilla Fine-Tuning. Suggesting that Wikidata-grounded feature graph might not provide a reliable average gain in the default schedule. By contrast, the LLM-direct ablation arm, $\textsc{LLM-LoRA-attn~FT}$, which uses the same adaptation pipeline but replaces Wikidata-grounded edges with the LLM's own proposed feature--feature graph, regresses the small-TFM mean relative to Vanilla FT. We attribute this to the lower density of the derived graphs as well as the 

%\label{sec:results:wikidata}

\textbf{General knowledge graphs might recover structure (Q3).} \Cref{tab:struct_kgwd} reveals that the gap between the ground truth structure graph and the Wikidata extracted graph is narrow for both tested models (the gap is $0.002$ for both variants). Regarding pooling, we find that models trained on graphs constructed on the 1-hop protocol perform better than (or equal to) direct and ancestor protocols, across both models. We also find that for NanoTabPFN, using the union of all three pooling strategies can slightly degrade performance (WD union: $0.701$ vs WD direct: $0.706$), whereas for NanoTabICL, the difference between variants is negligible. This demonstrates that both the matching mechanism and adjacency construction are sensitive components that can affect the performance. Moreover, we find that the random (density-matched) control on small datasets is a strong steering structure for TabPFN (outperforming both ancestor and union mappings). However, due to the synthetic nature of the benchmark, where instances were recovered by sampling a SCM, in a similar way that our priors are sampled, the difference might originate from the models recognising the structure from their pretraining~\cite{zhang2026mitra} (see \Cref{fig:struct_kg_vs_dag} in \Cref{app:kg-stats}). We nevertheless include this control because it provides an idea of what ground-truth structure can bring when no curated KG exists, and a soft upper bound for what Wikidata grounding can recover with $5$--$11$ features. \looseness-1

\textbf{Limitations.}
%Few limitations originate from the design choices of this work. 
\modelname requires tabular data populated with metadata, which may be unavailable for some tasks. Mapping to knowledge graphs (KGs) can be an ambiguous task, and this ambiguity may propagate to the tuning stage~\cite{llm_disimbiguate}. Next, the benefits of KG-informed adaptation depend heavily on the synthetic prior used during pretraining; if the pretraining prior already captures the relevant dependency structures, additional (external) KG priors may not yield further improvements. Additionally, tabular foundation model architectures do not all expose feature-level attention in the same manner (see \Cref{sec:implementation}). While some models provide a natural column-attention interface, others primarily attend over rows or support-query examples. Consequently, the optimal KG injection mechanism may differ across backbones. Finally, the selected training dynamic setup (number of steps, optimiser choice) might be suboptimal across different model families and sizes, potentially affecting performance, especially for the large models \cite{rubachev2025finetuningtabular}.

\begin{table}[t]
\centering\footnotesize
\caption{Structure-graph ablation on three Bayesian-network datasets (asia, cancer, sachs) with LLM-mapped Wikidata QID mappings. All methods use the same training; only the source of the feature-feature mask changes. \emph{DAG}: ground-truth Bayesian-network edges. \emph{WD}: Wikidata edges between resolved QIDs at four pooling strategies - \emph{direct} (Wikidata predicates), \emph{1-hop} (shared neighbour entity), \emph{ancestor} (shared subclass-of ancestor), \emph{union} (any of the three). \emph{Random graph}: density matched to the DAG.} %\textbf{Bold} = best }
\label{tab:struct_kgwd}
\begin{tabular}{@{}lcc@{}}
\toprule
Method & NanoTabICL (big, base) & NanoTabPFN (big, base) \\
\midrule
\multicolumn{3}{@{}l}{\textit{Baselines (no graph mask)}}\\
Pretrained & 0.510$_{\pm 0.018}$ & 0.502$_{\pm 0.006}$ \\
Vanilla FT & 0.697$_{\pm 0.148}$ & 0.700$_{\pm 0.149}$ \\
LoRA FT & 0.701$_{\pm 0.150}$ & 0.701$_{\pm 0.149}$ \\
\midrule
\multicolumn{3}{@{}l}{\textit{\modelname with ground-truth structure}}\\
\rowcolor{green!8} DAG (ground-truth) & \textbf{0.701$_{\pm 0.150}$} & \textbf{0.708$_{\pm 0.153}$} \\
\midrule
\multicolumn{3}{@{}l}{\textit{\modelname with Wikidata KG (varying pooling strategy)}}\\
\rowcolor{blue!6} WD direct & 0.697$_{\pm 0.149}$ & 0.706$_{\pm 0.153}$ \\
\rowcolor{blue!6} WD 1-hop & 0.699$_{\pm 0.149}$ & 0.706$_{\pm 0.152}$ \\
\rowcolor{blue!6} WD ancestor & 0.695$_{\pm 0.148}$ & 0.701$_{\pm 0.151}$ \\
\rowcolor{blue!6} WD union & 0.699$_{\pm 0.149}$ & 0.701$_{\pm 0.149}$ \\
\midrule
\multicolumn{3}{@{}l}{\textit{Density-matched control}}\\
\rowcolor{gray!10} Random graph & 0.694$_{\pm 0.148}$ & 0.705$_{\pm 0.153}$ \\
\bottomrule
\end{tabular}
\vspace{-3mm}
\end{table}

\section{Conclusion}

We have presented \modelname, a novel knowledge-informed framework for fine-tuning small foundation models. In specific domains, \modelname achieves substantial improvements over standard fine-tuning in both parameter efficiency and predictive performance. In more general settings, our results show that the approach's effectiveness depends on the task type and the knowledge-injection strategy. Overall, \modelname provides a reliable mechanism for incorporating structured knowledge when a ground-truth graph is available. We believe this opens new opportunities for developing small-scale, high-performance tabular foundation models. \looseness-1

\section*{Acknowledgments}
This work pas partly funded by the Slovenian Research Agency under the projects: P2-0103 and PR-12394. This work was also partially supported by the European Union’s Horizon Europe research and innovation program under grant agreement No. 101214398 (ELLIOT). Views and opinions expressed are, however, those of the author(s) only and do not necessarily reflect those of the European Union or the European Commission. Neither the European Union nor the European Commission can be held responsible for them. XJ acknowledges the generous support of the Google PhD Fellowship.
MJ and NS acknowledge the support of the U.S. Army Medical Research and Development Command of the Department of Defense; through the FY22 Breast Cancer Research Program of the Congressionally Directed Medical Research Programs, Clinical Research Extension Award GRANT13769713. Opinions, interpretations, conclusions, and recommendations are those of the authors and are not necessarily endorsed by the Department of Defense.

% For general domains, we proposed an agentic approach to constructing feature interaction graphs grounded in the Wikidata knowledge graph. While the gains were modest in comparison to domain-specific applications, they were partially influenced by the realistic and messy nature of general-domain data. We find that our method can reliably inject knowledge and perform meaningfully when a structured ground-truth graph exists.\looseness-1 

%\textbf{\color{blue}DON'T FORGET TO ADD THE CHECKLIST AT THE END}

%%%%%%%%%%%%%%%%%%%%%%%%%%%%%%%%%%%%%%%%%%%%%%
%\newpage
%\pagebreak
%\bibliographystyle{plainnat}
\bibliographystyle{plain}
\bibliography{references}

%%%%%%%%%%%%%%%%%%%%%%%%%%%%%%%%%%%%%%%%%%%%%%
%%%%%%%%%%%%%%%%%%%%%%%%%%%%%%%%%%%%%%%%%%%%%%
%%%%%%%%%%%%%%%%%%%%%%%%%%%%%%%%%%%%%%%%%%%%%%
%%%%%%%%%%%%%%%%%%%%%%%%%%%%%%%%%%%%%%%%%%%%%%
%%%%%%%%%%%%%%%%%%%%%%%%%%%%%%%%%%%%%%%%%%%%%%
%%%%%%%%%%%%%%%%%%%%%%%%%%%%%%%%%%%%%%%%%%%%%%
%%%%%%%%%%%%%%%%%%%%%%%%%%%%%%%%%%%%%%%%%%%%%%
%%%%%%%%%%%%%%%%%%%%%%%%%%%%%%%%%%%%%%%%%%%%%%
\newpage
\pagebreak
\appendix

%%% APPENDIX PRETRAINING

\section{Per-checkpoint pretraining configuration}
\label{app:pretrain_budget}

All 8 nanoscale variants of \Cref{tab:nanoscale_variants} are pretrained with an identical optimisation budget (see \Cref{tab:pretraining_budget}). \Cref{tab:prior_families} details synthetic prior families used for nanoscale pretraining.

\begin{table}[h]
\centering\footnotesize
\caption{Pretraining configuration per nano variant. Steps $\times$
Batch $=$ Tasks seen $= 12{,}000$ synthetic tabular tasks consumed during pretraining. Prior pool is the number of distinct synthetic tasks in the HDF5 prior dump. Passes = Tasks seen / Prior pool, i.e. how many times each prior task is seen on average (the loader wraps the pointer when it reaches the end of the file). Wallclock is per checkpoint on a single GPU. AdamW, learning rate $10^{-4}$, fixed seed 2402.}
\begin{tabular}{@{}llrrrrrr@{}}
\toprule
Backbone & Prior & Steps & Batch & Tasks seen & Prior pool & Passes & Wall (s) \\
\midrule
NanoTabICL  & \texttt{tabicl\_small} & 3{,}000 & 4 & 12{,}000 & 1{,}000 & 12 & 464--665 \\
NanoTabICL  & \texttt{tabicl\_big}   & 3{,}000 & 4 & 12{,}000 & 2{,}000 &  6 & 464--670 \\
NanoTabPFN  & \texttt{tabicl\_small} & 3{,}000 & 4 & 12{,}000 & 1{,}000 & 12 & 442--648 \\
NanoTabPFN  & \texttt{tabicl\_big}   & 3{,}000 & 4 & 12{,}000 & 2{,}000 &  6 & 446--654 \\
\bottomrule
\end{tabular}

\label{tab:pretraining_budget}
\end{table}

\begin{table}[h]
\caption{Nanoscale pretrained model variants. Each architecture-scale pair is
pretrained under both synthetic prior families, separating backbone, scale, and
prior-family effects.}
\label{tab:nanoscale_variants}
\centering
\small
\begin{tabular}{lllclc}
\toprule
Backbone & Prior family & Scale & Embedding & Blocks / layers & Parameters \\
\midrule
\multirow{4}{*}{NanoTabICL}
& \multirow{2}{*}{\texttt{tabicl\_small}}
& small & 48 & col/row/ICL: 2/2/2; heads: 4/4/4 & 848{,}770 \\
& & base  & 80 & col/row/ICL: 3/3/3; heads: 4/4/4 & 3{,}302{,}742 \\
\cmidrule(lr){2-6}
& \multirow{2}{*}{\texttt{tabicl\_big}}
& small & 48 & col/row/ICL: 2/2/2; heads: 4/4/4 & 848{,}770 \\
& & base  & 80 & col/row/ICL: 3/3/3; heads: 4/4/4 & 3{,}302{,}742 \\
\midrule
\multirow{4}{*}{NanoTabPFN}
& \multirow{2}{*}{\texttt{tabicl\_small}}
& small & 128 & layers: 4; heads: 4; MLP: 512 & 1{,}129{,}994 \\
& & base  & 192 & layers: 6; heads: 6; MLP: 768 & 3{,}717{,}514 \\
\cmidrule(lr){2-6}
& \multirow{2}{*}{\texttt{tabicl\_big}}
& small & 128 & layers: 4; heads: 4; MLP: 512 & 1{,}129{,}994 \\
& & base  & 192 & layers: 6; heads: 6; MLP: 768 & 3{,}717{,}514 \\
\bottomrule
\end{tabular}
\end{table}

\begin{table}[h]
\caption{Synthetic prior families used for nanoscale pretraining. Both priors
come from the same SCM-based mixed generator family, but differ in feature range
and amount of generated prior data.}
\label{tab:prior_families}
\centering
\small
\begin{tabular}{lccccc}
\toprule
Prior family & Size & Batches & Feature range & Max length & Max classes \\
\midrule
\texttt{tabicl\_small} & 148 MB & 1000 & $5$--$60$ & 300 & 8 \\
\texttt{tabicl\_big}   & 1.1 GB & 2000 & $40$--$200$ & 300 & 8 \\
\bottomrule
\end{tabular}
\end{table}

%%% APPENDIX PSEUDO
\section{\modelname-Algorithm}
\label{app:kglora}
In this appendix, we outline the exact algorithm for knowledge adaptation of the TFMs.

\begin{algorithm}[H]
\caption{KG-LoRa fine-tuning of a small tabular foundation model. Lines~5--12
attach the KG-derived structural prior $S^{(\rho)}$ as per-block additive logit
slots; lines~13--14 attach LoRA factors. Only $\Phi$ -- the LoRA factors $\{L_A^{(\ell,*)},
L_B^{(\ell,*)}\}$ and the soft-slot scalars $\{\beta^{(\ell)}\}_{\sigma_\ell=\textsc{soft}}$
 is updated; the pretrained weights $\theta$ and the structural prior $S^{(\rho)}$
remain frozen.}
\label{alg:kg_lora_attn}
\begin{algorithmic}[1]
\Require pretrained TFM $f_\theta$ with $L$ feature-attention blocks of $H$ heads
\Require dataset $(X,y)$ with feature names $c_{1{:}F}$, knowledge graph $\mathcal{KG}=(V,E)$
\Require schedule $\boldsymbol{\sigma}=(\sigma_1,\dots,\sigma_L)\in\{\textsc{hard},\textsc{soft},\textsc{off}\}^L$
\Require LoRA rank $r$, scale $\alpha$; learning rate $\eta$; FT steps $T$
\Statex \textbf{1.~~Project the KG onto the feature axis} \hfill \textit{Section \ref{sec:kg_mapping})}
\State for $f \in c_{1{:}F}$:\quad $q_f \gets$ \Call{MapToEntity}{$f,\mathcal{KG}$} \Comment{biomedical: PrimeKG; general: agentic Wikidata; $\varnothing$ if unmapped}
\State $S^{(\rho)}_{ij} \gets \mathbb{1}[\,(q_i,*,q_j)\in E\text{ within }\rho\text{ hops},\;q_i,q_j\ne\varnothing\,]$
\Statex \textbf{2.~~Build per-block injection slots} \hfill \textit{(Section~\ref{sec:method_injection})}
\State $\Phi \gets \emptyset$ \Comment{trainable parameters}
\For{$\ell = 1,\dots,L$}
    \If{$\sigma_\ell = \textsc{off}$} \State \textbf{continue}
    \ElsIf{$\sigma_\ell = \textsc{hard}$}
        \State $M^{(\ell)}_{ij} \gets 0$ if $S^{(\rho)}_{ij}{=}1$ else $-\infty$ \Comment{shared across heads, no params}
    \ElsIf{$\sigma_\ell = \textsc{soft}$}
        \State init $\beta^{(\ell)}\!\in\!\mathbb{R}^H \gets \mathbf{0.5}$;\quad $\Phi\gets\Phi\cup\{\beta^{(\ell)}\}$
        \State $M^{(\ell,h)}_{ij} \gets \beta^{(\ell)}_h \cdot S^{(\rho)}_{ij}$ \Comment{differentiable, per-head gate}
    \EndIf
\EndFor
\Statex \textbf{3.~~Attach LoRA on every linear projection} \hfill \textit{(QKV, out\_proj, FFN)}
\State for each frozen projection $W$: init $L_A\!\sim\!\mathcal{N}(0,1/r)$, $L_B\!\gets\!0$;\quad $W_{\text{eff}}\gets W+\tfrac{\alpha}{r}L_BL_A$
\State $\Phi \gets \Phi \cup \{L_A^{(\ell,*)},L_B^{(\ell,*)} : \ell\in[L]\}$
\Statex \textbf{4.~~Episodic in-context fine-tune} \hfill \textit{(Section \ref{sec:experimental_setup})}
\For{$t = 1, \dots, T$}
    \State sample stratified split $(\mathcal{S}_c,\mathcal{S}_q)$ of $(X,y)$
    \For{block $\ell\in[L]$, head $h\in[H]$ in feature attention}
        \State $Z^{(\ell,h)} \gets Q^{(\ell,h)}K^{(\ell,h)\top}/\sqrt{d_h} + M^{(\ell,h)}$ \Comment{$M^{(\ell,h)}{=}0$ if $\sigma_\ell{=}\textsc{off}$}
    \EndFor
    \State $\hat{y}_q \gets f_\theta(X_{\mathcal{S}_c},y_{\mathcal{S}_c},X_{\mathcal{S}_q})$ \Comment{forward through KG-aware $f_\theta$}
    \State $\mathcal{L} \gets \mathrm{CE}(\hat{y}_q, y_{\mathcal{S}_q})$
    %\State $\Phi \gets \Phi - \eta\,\nabla_\Phi\mathcal{L}$ \Comment{$\theta$ and $S^{(\rho)}$ are frozen}
    \State $\Phi \gets \Call{AdamW}{\Phi,\eta\nabla_\Phi\mathcal{L}}$ \Comment{$\theta$ and $S^{(\rho)}$ are frozen}
\EndFor
\State \Return adapted $f_\theta$ with $W_{\text{eff}}$ and active slots $\{M^{(\ell)}\}$
\end{algorithmic}
\end{algorithm}

%%% APPENDIX DATASETS

\section{Dataset selection and metadata}
\label{app:dataset_meta}

This appendix documents the inclusion criteria and resulting metadata for the three dataset families used in \Cref{sec:experimental_setup}. 

\subsection{Biomedical: CUMIDA-42}

CUMIDA~\citep{cumida2019} ships 69 cancer microarray datasets in its public catalog. We retain the 42 panels that satisfy all of the following criteria, applied programmatically and audited per dataset: (i) a non-empty symbol-mapped expression matrix exists, (ii) the target defines a non-trivial classification task with at least two classes and enough minority-class samples for stratified 3-fold cross-validation, and (iii) the panel contains at least $N \ge 12$ patient samples. We do not otherwise filter on sample count; the smallest retained panel is \texttt{Breast\_GSE26910} with $N=12$, and the largest is \texttt{Leukemia\_GSE28497} with $N=281$. \Cref{tab:cumida42_meta} gives the resulting CUMIDA-42 suite. ``$N$'' is the number of patient samples, ``$K_\text{full}$'' is the number of symbol-mapped expression features before fold-local selection, ``cls'' is the number of outcome classes, and ``maj.\,\%'' is the majority-class share.

\begin{table}[h]
\caption{
The 42 CUMIDA cancer panels used in \Cref{tab:cumida42_no_rnd}.
Source: \url{https://sbcb.inf.ufrgs.br/cumida}.
}
\label{tab:cumida42_meta}
\centering
\footnotesize
\begin{tabular}{@{}llrrrr@{}}
\toprule
\# & Dataset & Cancer & $N$ & $K_\text{full}$ & cls (maj.\,\%) \\
\midrule
 1 & \texttt{Bladder\_GSE31189}           & Bladder    &  85 & 20{,}779 & 2 (56\%) \\
 2 & \texttt{Brain\_GSE15824}             & Brain      &  37 & 20{,}779 & 4 (32\%) \\
 3 & \texttt{Brain\_GSE50161}             & Brain      & 108 & 20{,}779 & 4 (43\%) \\
 4 & \texttt{Breast\_GSE10797}            & Breast     &  66 & 12{,}939 & 3 (42\%) \\
 5 & \texttt{Breast\_GSE26910}            & Breast     &  12 & 20{,}779 & 2 (50\%) \\
 6 & \texttt{Breast\_GSE42568}            & Breast     & 116 & 20{,}779 & 2 (87\%) \\
 7 & \texttt{Breast\_GSE45827}            & Breast     & 151 & 20{,}779 & 6 (27\%) \\
 8 & \texttt{Breast\_GSE7904}             & Breast     &  45 & 20{,}779 & 3 (47\%) \\
 9 & \texttt{Colorectal\_GSE21510}        & Colorectal & 147 & 20{,}779 & 3 (71\%) \\
10 & \texttt{Colorectal\_GSE32323}        & Colorectal &  33 & 20{,}779 & 2 (52\%) \\
11 & \texttt{Colorectal\_GSE41328}        & Colorectal &  18 & 20{,}779 & 2 (56\%) \\
12 & \texttt{Colorectal\_GSE44861}        & Colorectal & 105 & 12{,}939 & 2 (50\%) \\
13 & \texttt{Colorectal\_GSE77953}        & Colorectal &  55 & 12{,}939 & 4 (31\%) \\
14 & \texttt{Colorectal\_GSE8671}         & Colorectal &  63 & 20{,}779 & 2 (51\%) \\
15 & \texttt{Gastric\_GSE19826}           & Gastric    &  24 & 20{,}779 & 2 (50\%) \\
16 & \texttt{Gastric\_GSE79973}           & Gastric    &  20 & 20{,}779 & 2 (50\%) \\
17 & \texttt{Leukemia\_GSE14317}          & Leukemia   &  25 & 12{,}939 & 2 (72\%) \\
18 & \texttt{Leukemia\_GSE22529\_U133A}   & Leukemia   &  52 & 12{,}939 & 2 (79\%) \\
19 & \texttt{Leukemia\_GSE22529\_U133B}   & Leukemia   &  52 & 10{,}184 & 2 (79\%) \\
20 & \texttt{Leukemia\_GSE28497}          & Leukemia   & 281 & 12{,}939 & 7 (26\%) \\
21 & \texttt{Leukemia\_GSE63270}          & Leukemia   & 101 & 20{,}779 & 2 (59\%) \\
22 & \texttt{Leukemia\_GSE71935}          & Leukemia   &  46 & 20{,}779 & 2 (80\%) \\
23 & \texttt{Leukemia\_GSE9476}           & Leukemia   &  64 & 12{,}939 & 5 (41\%) \\
24 & \texttt{Liver\_GSE14520\_U133\_2}    & Liver      &  41 & 12{,}939 & 2 (54\%) \\
25 & \texttt{Liver\_GSE22405}             & Liver      &  48 & 12{,}939 & 2 (50\%) \\
26 & \texttt{Liver\_GSE60502}             & Liver      &  36 & 12{,}939 & 2 (50\%) \\
27 & \texttt{Liver\_GSE62232}             & Liver      &  91 & 20{,}779 & 2 (89\%) \\
28 & \texttt{Lung\_GSE18842}              & Lung       &  90 & 20{,}779 & 2 (51\%) \\
29 & \texttt{Lung\_GSE19804}              & Lung       & 114 & 20{,}779 & 2 (51\%) \\
30 & \texttt{Lung\_GSE27262}              & Lung       &  48 & 20{,}779 & 2 (50\%) \\
31 & \texttt{Lung\_GSE7670}               & Lung       &  51 & 12{,}939 & 2 (53\%) \\
32 & \texttt{Ovary\_GSE6008}              & Ovary      &  98 & 12{,}939 & 4 (42\%) \\
33 & \texttt{Pancreatic\_GSE16515}        & Pancreatic &  51 & 20{,}779 & 2 (71\%) \\
34 & \texttt{Prostate\_GSE26910}          & Prostate   &  12 & 20{,}779 & 2 (50\%) \\
35 & \texttt{Prostate\_GSE46602}          & Prostate   &  49 & 20{,}779 & 2 (71\%) \\
36 & \texttt{Prostate\_GSE55945}          & Prostate   &  17 & 20{,}779 & 2 (59\%) \\
37 & \texttt{Prostate\_GSE6919\_U95Av2}   & Prostate   & 124 &  9{,}013 & 2 (50\%) \\
38 & \texttt{Prostate\_GSE6919\_U95B}     & Prostate   & 124 &  6{,}820 & 2 (52\%) \\
39 & \texttt{Renal\_GSE53757}             & Renal      & 143 & 20{,}779 & 2 (50\%) \\
40 & \texttt{Renal\_GSE6344\_U133A}       & Renal      &  20 & 12{,}939 & 2 (50\%) \\
41 & \texttt{Renal\_GSE6344\_U133B}       & Renal      &  20 & 10{,}184 & 2 (50\%) \\
42 & \texttt{Renal\_GSE66270}             & Renal      &  28 & 20{,}779 & 2 (50\%) \\
\bottomrule
\end{tabular}
\end{table}

Sample feature names (first ten gene symbols of
\texttt{Leukemia\_GSE9476} after probe-to-symbol normalisation, before
fold-local RF feature selection):
\texttt{TP53}, \texttt{BRCA1}, \texttt{KMT2A}, \texttt{RUNX1},
\texttt{FLT3}, \texttt{NPM1}, \texttt{IDH1}, \texttt{IDH2},
\texttt{DNMT3A}, \texttt{TET2} (illustrative; the actual selected gene set is fold-specific).

\textbf{Feature pre-processing.}
CUMIDA panels contain between 6{,}820 and 20{,}779 symbol-mapped expression features, which exceeds the synthetic-prior feature range of our nanoscale checkpoints, $K \in [40,200]$. We therefore apply fold-local feature selection. On each training fold, we fit a Random Forest classifier~\cite{rf} with 100 trees and maximum depth 10, select the top-$T$ features by Gini importance, and add $R$ disjoint random features, yielding $K=T+R$. The headline arm uses $T=50,R=50$; \Cref{sec:feat_arm} reports sensitivity sweeps for $\{T=100,R=0\}$, $\{T=50,R=50\}$, and $\{T=25,R=75\}$.

\subsection{Cross-domain UCI/OpenML (general-knowledge)}

The 13 datasets in \Cref{tab:xdomain13_meta} were selected by applying three constraints: (i) each dataset must provide feature names or descriptions with enough natural-language semantics for the agentic Wikidata mapper of \Cref{sec:kg_mapping}; (ii) each dataset must have $K \le 50$ raw features, so all columns can enter $S^{(\rho)}$ without fold-local feature selection; and (iii) each dataset must contain at least $N \ge 150$ samples and at least five samples in each minority class. The retained suite spans $8 \le K \le 41$, from \texttt{hepatitis} with $N=155$ to \texttt{stroke} with $N=4{,}909$.

We do not filter on Wikidata-mapping coverage before running experiments.
Low-coverage datasets such as \texttt{QSAR-biodeg} and \texttt{dermatology} are retained deliberately, since these failures are part of the empirical setting studied in \Cref{tab:kg_llm_stats} in \Cref{app:kg-stats}.

\begin{table}[h]
\centering
\footnotesize
\caption{
The 13 cross-domain UCI/OpenML datasets used in
\Cref{tab:cross_domain_13ds}. ``$N$'' = total samples, ``$K$'' =
features, ``cls'' = classes, ``maj.\,\%'' = majority-class share.
The agentic v2 Wikidata mapper of \Cref{sec:kg_mapping} produces
the $S^{(\rho)}$ adjacency for each dataset; mapping coverage and edge
statistics are reported in \Cref{tab:kg_llm_stats} in \Cref{app:kg-stats}.
}
\begin{tabular}{@{}lrrrr@{}}
\toprule
Dataset & $N$ & $K$ & cls & maj.\,\% \\
\midrule
\texttt{automobile}         &  159 & 25 & 2 & 73.6 \\
\texttt{cirrhosis}          &  418 & 17 & 3 & 55.5 \\
\texttt{dermatology}        &  366 & 34 & 6 & 30.6 \\
\texttt{diabetes\_pima}     &  768 &  8 & 2 & 65.1 \\
\texttt{glass}              &  214 &  9 & 6 & 35.5 \\
\texttt{glioma}             &  839 & 23 & 2 & 58.0 \\
\texttt{heart-h}            &  294 & 13 & 2 & 63.9 \\
\texttt{heart-statlog}      &  270 & 13 & 2 & 55.6 \\
\texttt{hepatitis}          &  155 & 19 & 2 & 79.4 \\
\texttt{QSAR-biodeg}        & 1055 & 41 & 2 & 66.3 \\
\texttt{SPECT}              &  267 & 22 & 2 & 79.4 \\
\texttt{steel-plates-fault} & 1941 & 27 & 7 & 34.7 \\
\texttt{stroke}             & 4909 & 10 & 2 & 95.7 \\
\bottomrule
\end{tabular}

\label{tab:xdomain13_meta}
\end{table}

Sample feature names (illustrative, three datasets):

\paragraph{\texttt{glass}.}
\texttt{RI} (refractive index), \texttt{Na} (sodium),
\texttt{Mg} (magnesium), \texttt{Al} (aluminum), \texttt{Si} (silicon),
\texttt{K} (potassium), \texttt{Ca} (calcium), \texttt{Ba} (barium),
\texttt{Fe} (iron).
\paragraph{Feature pre-processing.}
All $K$ columns are used verbatim. We label-encode categorical columns and standardise numeric columns using training-fold statistics only. For datasets with more than 1{,}000 samples, namely \texttt{stroke}, \texttt{QSAR-biodeg}, and \texttt{steel-plates-fault}, we use a fixed-seed stratified subsample of 1{,}000 examples, shared across all method variants.

%%% APPENDIX LLM

\section{LLM prompts and the LLM's role in each method variant}
\label{app:llm_prompts}

This appendix documents the exact prompts used by the agentic Wikidata mapper  and the LLM-direct graph (\Cref{sec:kg_mapping}), and clarifies the LLM's role in the random-graph control of \Cref{tab:struct_kgwd}. All LLM calls use \texttt{gemini-3-pro-preview} with \texttt{temperature=0.0} and a JSON-only response format.

\subsection{Agentic Wikidata mapper, Stage 1: query proposal}

A single batched LLM call sees the dataset description and all column descriptions at once. The LLM proposes Wikidata search queries; it does not itself emit QIDs at this stage.

\begin{quote}
\footnotesize\ttfamily
DATASET: \{dataset\_desc\}\\[2pt]

TASK: For each column below, propose 3--7 Wikidata search queries that retrieve the canonical Wikidata entity for the column's *concept*. Use synonyms, expanded abbreviations, and parent-concept terms. Use cross-column context (e.g., seeing other clinical features helps you expand `trestbps' to `resting blood pressure').\\[2pt]

Mark `value\_level: true' if the *values* of the column are entities
(e.g., a `cap-shape' column in a mushroom dataset whose values are
`convex', `flat', `knobbed') rather than the column itself.\\[2pt]

Provide a `type\_hint' from this set:
gene $\mid$ chemical $\mid$ disease $\mid$ measurement $\mid$ anatomy
$\mid$ clinical\_finding $\mid$ categorical $\mid$ demographic
$\mid$ physical\_property $\mid$ other.\\[2pt]

COLUMNS: \{cols\_payload\}\\[2pt]

Respond with strict JSON, one entry per column:\\
\{"<col\_name>": \{"queries": ["q1", "q2", ...], "type\_hint": "...",
"value\_level": <bool>, "rationale": "..."\}\}
\end{quote}

\subsection{Agentic Wikidata mapper, Stage 3: disambiguation with abstention}

A second batched LLM call sees, for each column, the candidate list
returned by the Wikidata REST API (\texttt{wbsearchentities}) plus the
dataset context. The LLM either selects one QID from the candidate list
or abstains. This stage's abstention discipline is the load-bearing
difference from a naive ``prompt-the-LLM-for-the-QID'' baseline
%(see Appendix~\ref{app:proflash} for the Pro-vs-Flash quality screen).

\begin{quote}
\footnotesize\ttfamily
DATASET: \{dataset\_desc\}\\[2pt]

TASK: For each column, pick the BEST Wikidata QID from its candidate
list, or return empty string if none fit. Prefer concrete concepts over
scientific articles. Use cross-column context. Refuse to invent QIDs not
in the candidate list.\\[2pt]

COLUMNS\_AND\_CANDIDATES: \{payload\}\\[2pt]

Respond with strict JSON:\\
\{"<col\_name>": \{"qid": "<Q123 or empty>", "rationale": "...",
"confidence": 0.0--1.0\}\}.
\end{quote}

Stages 2 (Wikidata candidate retrieval) and 4 (SPARQL edge discovery) are
deterministic and do not involve the LLM.

\subsection{LLM-direct graph (negative control)}

A single LLM call proposes feature--feature edges directly from the
dataset and column descriptions, with no Wikidata lookup. The LLM emits
edges with a self-reported strength in $[0,1]$ and a one-sentence
rationale; we keep edges with strength $\geq 0.5$ and binarise.

\begin{quote}
\footnotesize\ttfamily
DATASET: \{dataset\_desc\}\\[2pt]

COLUMNS: \{cols\_payload\}\\[2pt]

TASK: Propose semantically meaningful feature-feature edges that capture
*useful structural priors* for an attention mechanism. Two columns
should be connected if their interaction is plausibly informative for
predicting the target --- for example: same biological pathway, same
body system, same chemical class, same physical mechanism, known
causal/correlational relationship in the literature.\\[2pt]

Use your *implicit domain knowledge* from training. Do NOT just connect
syntactically similar names --- connect features that are
semantically/causally related.\\[2pt]

Each edge should include a one-sentence rationale and a strength score
in [0.5, 1.0] (0.5 = weakly related, 1.0 = strongly/canonically related).\\[2pt]

Be selective: aim for 1--3 edges per column on average, prioritising
strong edges. Symmetric undirected edges, no self-loops.\\[2pt]

Respond with strict JSON:\\
\{"edges": [\{"from": "<col\_a>", "to": "<col\_b>",
"relation": "<short label>", "rationale": "...", "strength": <float>\}, ...],
"global\_notes": "<1-2 sentences on the overall structure you propose>"\}
\end{quote}

%\subsection{The LLM's role in the random-graph control}
The $\textsc{Random graph}$ arm of \Cref{tab:struct_kgwd} uses a density-matched random graph $\widetilde{S}$ generated by a fixed NumPy random seed: given the target density $\mathrm{dens}(S^{(\rho)})$ of the corresponding Wikidata or DAG adjacency, $\widetilde{S}_{ij}$ is drawn i.i.d.\ Bernoulli with probability matched to that density, symmetrised, with self-loops added. There is no LLM call, no Wikidata lookup, and no semantic content. The control therefore isolates whether
attention-mask sparsity \emph{alone} is what the schedule consumes, independent of the meaning of the masked entries.

%%% IMPLEMENTATION APPENDIX
\section{Implementation}
\label{sec:implementation}

Our implementation builds KG-LoRA-attn on top of two inspectable nano tabular backbones: NanoTabPFN (\url{https://github.com/automl/nanoTabPFN}) and the minimal NanoTabICL/TabICL-style code linked from the TabICL v2 repository (\url{https://github.com/soda-inria/nanotabicl}). The nano checkpoints are pretrained from explicit synthetic prior dumps; we follow the prior-dump interface used in the nanoTabPFN/TFM-Playground ecosystem (\url{https://github.com/automl/TFM-Playground/}), and treat the prior family as part of the model identity. For data analysis and preprocessing we use Scikit-Learn~\cite{scikit-learn}, for neural training we use PyTorch~\cite{paszke2019pytorch}. For each downstream fold, feature selection, scaling, and KG construction are fit only on the training split. The selected features define a frozen graph prior $S^{(\rho)}$, built from PrimeKG for CUMIDA gene-expression panels and from Wikidata mappings for general-domain tables. In each adapted feature-attention block, the prior is injected only as a pre-softmax logit term. The default schedule is $(\textsc{hard},\textsc{soft},\textsc{soft})$: the first adapted block applies a hard off-graph mask, while later blocks use a learnable per-head soft bias $g_h^{(\ell)}S^{(\rho)}$ with $g_h^{(\ell)}=\sigma(a_h^{(\ell)})s_h^{(\ell)}$, initialized to $0.5$. Task adaptation uses LoRA with rank $r=16$ and $\alpha=32$; for the nano backbones we attach LoRA to QKV, attention output, and feed-forward projections. The public TabICL v2 wrapper keeps QKV-LoRA disabled because its upstream attention module needs additional signature and mask-handling fixes, so that wrapper uses LoRA on output and MLP projections only. Experiments are run in an Apptainer container (\url{https://apptainer.org/documentation/}). Statistical comparisons use paired tests and AutoRank (\url{https://github.com/sherbold/autorank}; \cite{Herbold2020}).

\subsection{Hyperparameters}
\label{sec:hyperparameters}

Unless otherwise stated, downstream runs use 500 fine-tuning steps, AdamW~\cite{loshchilov2018decoupled} optimization as implemented in the release trainer, learning rate $10^{-4}$ for full-model Vanilla FT, learning rate $10^{-3}$ for LoRA and \modelname, LoRA rank $r=16$, LoRA scale $\alpha=32$, and the schedule $(\textsc{hard},\textsc{soft},\textsc{soft})$. CUMIDA uses the \texttt{rf50\_rnd50} feature arm by default, with additional feature-arm tables for \texttt{rf100}, \texttt{rf25\_rnd75}, and random-only selection. The xdomain13 HPO columns are validation-selected sweeps over the two big-prior base nano backbones: Vanilla FT sweeps $\eta_{\mathrm{full}}\in\{10^{-5},5{\times}10^{-5},10^{-4},5{\times}10^{-4},10^{-3}\}$, while KG-LoRA-attn sweeps $r\in\{8,16,32\}$ with $\alpha=2r$, $\eta_{\mathrm{LoRA}}\in\{5{\times}10^{-4},10^{-3},2{\times}10^{-3}\}$, and all eight schedules in $\{\textsc{hard},\textsc{soft}\}^3$. Reported default tables use 3 seeds $\times$ 3 folds when available; release smoke tests may use smaller settings for runtime.

%% CUMOIDA ALL DATA 
\section{Full CUMIDA-42 results across feature-selection arms}
\label{sec:feat_arm}

This appendix reports the full per-checkpoint CUMIDA-42 result tables for all four feature-selection arms in increasing order of input noise:

\begin{itemize}
\item \texttt{top100} --- top100 RF-importance columns (cleanest signal).
\item \texttt{top50\_rnd50} --- 50 RF-top + 50 disjoint random columns (\textbf{headline} of the main paper).
\item \texttt{top25\_rnd75} --- 25 RF-top + 75 disjoint random columns (noisier than headline).
\item \texttt{rnd100} --- 100 uniformly-sampled columns (no RF supervision in feature selection).
\end{itemize}

Across all four arms, each cell reports the mean balanced accuracy over the $3 \text{ seeds} \times 3 \text{ folds} = 9$ splits for each $(\text{checkpoint}, \text{dataset})$ pair, followed by averaging across the 42 CUMIDA panels. `$\Delta$ vs Vanilla FT'' is computed by aggregating the per-cell differences at the $(\text{checkpoint}, \text{dataset})$ level. Foundation-model ceilings (TabICL~v2, 27.05\,M; TabPFN~v2.6, 10.73\,M) are evaluated using the same protocol. The v2 fine-tuning cells are reported for the headline \texttt{top50\_rnd50} and \texttt{rnd100} arms (\Cref{tab:cumida42_top50rnd50} and \Cref{tab:cumida42_random100}); the corresponding v2 fine-tuning cells for the cleaner \texttt{top100} and noisier \texttt{top25\_rnd75}.

\begin{table}[H]
\centering\footnotesize
\caption{Mean balanced accuracy on the CUMIDA-42 gene-expression benchmark, with the \emph{RF-100} feature arm (top100 RF-importance columns; cleanest signal). \textbf{Bold} = best in row. \emph{Zero-Shot}: zero-shot in-context inference. \emph{Vanilla FT}: full-model episodic fine-tuning. \emph{LoRA FT}: tuning without graph mask. \emph{\modelname}: our proposed method.}
\label{tab:cumida42_top100}
\resizebox{\textwidth}{!}{
    \begin{tabular}{@{}lrcccc@{}}
    \toprule
    Model & Params & Zero-Shot & Vanilla FT & LoRA FT & \modelname \\
    \midrule
    NanoTabICL big-prior, base & 3.30 M & 0.603$_{\pm 0.262}$ & 0.797$_{\pm 0.176}$ & 0.808$_{\pm 0.170}$ & \textbf{0.809$_{\pm 0.170}$} \\
    NanoTabICL big-prior, small & 0.85 M & 0.544$_{\pm 0.227}$ & 0.778$_{\pm 0.186}$ & \textbf{0.784$_{\pm 0.181}$} & 0.778$_{\pm 0.182}$ \\
    NanoTabICL small-prior, base & 3.30 M & 0.662$_{\pm 0.262}$ & 0.810$_{\pm 0.175}$ & 0.826$_{\pm 0.164}$ & \textbf{0.829$_{\pm 0.162}$} \\
    NanoTabICL small-prior, small & 0.85 M & 0.498$_{\pm 0.182}$ & 0.772$_{\pm 0.179}$ & 0.789$_{\pm 0.182}$ & \textbf{0.793$_{\pm 0.177}$} \\
    NanoTabPFN big-prior, base & 3.72 M & 0.427$_{\pm 0.137}$ & 0.747$_{\pm 0.187}$ & 0.744$_{\pm 0.192}$ & \textbf{0.766$_{\pm 0.188}$} \\
    NanoTabPFN big-prior, small & 1.13 M & 0.439$_{\pm 0.130}$ & 0.736$_{\pm 0.188}$ & 0.747$_{\pm 0.184}$ & \textbf{0.774$_{\pm 0.185}$} \\
    NanoTabPFN small-prior, base & 3.72 M & 0.446$_{\pm 0.107}$ & 0.756$_{\pm 0.189}$ & 0.752$_{\pm 0.189}$ & \textbf{0.774$_{\pm 0.179}$} \\
    NanoTabPFN small-prior, small & 1.13 M & 0.451$_{\pm 0.119}$ & 0.740$_{\pm 0.193}$ & 0.758$_{\pm 0.185}$ & \textbf{0.784$_{\pm 0.187}$} \\
    \midrule
    \rowcolor{gray!12} \textit{mean over 8 nano ckpts} & --- & 0.509$_{\pm 0.205}$ & 0.767$_{\pm 0.186}$ & 0.776$_{\pm 0.183}$ & \textbf{0.788$_{\pm 0.180}$} \\
    \rowcolor{gray!12} \textit{$\Delta$ vs Vanilla FT} & --- & -0.259 & --- & +0.009 & \textbf{+0.021} \\
    \midrule
    \rowcolor{blue!8} TabICL v2 & 27.05 M & \textbf{0.905$_{\pm 0.119}$} & --- & --- & --- \\
    \rowcolor{blue!8} TabPFN v2.6 & 10.73 M & \textbf{0.903$_{\pm 0.127}$} & --- & --- & --- \\
    \bottomrule
    \end{tabular}
}
\end{table}

\begin{table}[H]
\centering\footnotesize
\caption{Mean balanced accuracy on the CUMIDA-42 gene-expression benchmark, with the \emph{RF-50 + 50 random} feature arm (50 top-RF + 50 disjoint random columns; \textbf{headline}). \textbf{Bold} = best in row. \emph{Zero-Shot}: zero-shot in-context inference. \emph{Vanilla FT}: full-model episodic fine-tuning. \emph{LoRA FT}: tuning without graph mask. \emph{\modelname}: our proposed method.}
\label{tab:cumida42_top50rnd50}
\resizebox{\textwidth}{!}{
    \begin{tabular}{@{}lrcccc@{}}
    \toprule
    Model & Params & Zero-Shot & Vanilla FT & LoRA FT & \modelname \\
    \midrule
    NanoTabICL big-prior, base & 3.30 M & 0.582$_{\pm 0.239}$ & 0.772$_{\pm 0.145}$ & \textbf{0.796$_{\pm 0.141}$} & 0.793$_{\pm 0.139}$ \\
    NanoTabICL big-prior, small & 0.85 M & 0.543$_{\pm 0.210}$ & 0.748$_{\pm 0.141}$ & 0.758$_{\pm 0.137}$ & \textbf{0.767$_{\pm 0.145}$} \\
    NanoTabICL small-prior, base & 3.30 M & 0.649$_{\pm 0.257}$ & 0.792$_{\pm 0.140}$ & 0.808$_{\pm 0.142}$ & \textbf{0.815$_{\pm 0.137}$} \\
    NanoTabICL small-prior, small & 0.85 M & 0.475$_{\pm 0.145}$ & 0.751$_{\pm 0.148}$ & 0.767$_{\pm 0.140}$ & \textbf{0.768$_{\pm 0.151}$} \\
    NanoTabPFN big-prior, base & 3.72 M & 0.427$_{\pm 0.118}$ & 0.733$_{\pm 0.138}$ & 0.723$_{\pm 0.140}$ & \textbf{0.747$_{\pm 0.135}$} \\
    NanoTabPFN big-prior, small & 1.13 M & 0.441$_{\pm 0.117}$ & 0.706$_{\pm 0.137}$ & 0.719$_{\pm 0.133}$ & \textbf{0.757$_{\pm 0.150}$} \\
    NanoTabPFN small-prior, base & 3.72 M & 0.443$_{\pm 0.107}$ & 0.724$_{\pm 0.145}$ & 0.720$_{\pm 0.139}$ & \textbf{0.733$_{\pm 0.148}$} \\
    NanoTabPFN small-prior, small & 1.13 M & 0.448$_{\pm 0.110}$ & 0.704$_{\pm 0.144}$ & 0.720$_{\pm 0.145}$ & \textbf{0.767$_{\pm 0.139}$} \\
    \midrule
    \rowcolor{gray!12} \textit{mean over 8 nano ckpts} & --- & 0.501$_{\pm 0.187}$ & 0.741$_{\pm 0.144}$ & 0.751$_{\pm 0.142}$ & \textbf{0.768$_{\pm 0.144}$} \\
    \rowcolor{gray!12} \textit{$\Delta$ vs Vanilla FT} & --- & -0.240 & --- & +0.010 & \textbf{+0.027} \\
    \midrule
    \rowcolor{blue!8} TabICL v2 & 27.05 M & \textbf{0.893$_{\pm 0.114}$} & 0.884$_{\pm 0.116}$ & 0.884$_{\pm 0.116}$ & 0.876$_{\pm 0.120}$ \\
    \rowcolor{blue!8} TabPFN v2.6 & 10.73 M & \textbf{0.892$_{\pm 0.118}$} & 0.862$_{\pm 0.127}$ & 0.857$_{\pm 0.128}$ & 0.865$_{\pm 0.121}$ \\
    \bottomrule
    \end{tabular}
}
\end{table}

\begin{table}[H]
\centering\footnotesize
\caption{Mean balanced accuracy on the CUMIDA-42 gene-expression benchmark, with the \emph{RF-25 + 75 random} feature arm (25 top-RF + 75 disjoint random columns; noisier than headline). \textbf{Bold} = best in row. \emph{Zero-Shot}: zero-shot in-context inference. \emph{Vanilla FT}: full-model episodic fine-tuning. \emph{LoRA FT}: tuning without graph mask. \emph{\modelname}: our proposed method.}
\label{tab:cumida42_top2575}
\resizebox{\textwidth}{!}{
    \begin{tabular}{@{}lrcccc@{}}
    \toprule
    Model & Params & Zero-Shot & Vanilla FT & LoRA FT & \modelname \\
    \midrule
    NanoTabICL big-prior, base & 3.30 M & 0.557$_{\pm 0.237}$ & 0.745$_{\pm 0.186}$ & \textbf{0.764$_{\pm 0.183}$} & 0.762$_{\pm 0.188}$ \\
    NanoTabICL big-prior, small & 0.85 M & 0.522$_{\pm 0.206}$ & 0.713$_{\pm 0.195}$ & 0.732$_{\pm 0.186}$ & \textbf{0.741$_{\pm 0.189}$} \\
    NanoTabICL small-prior, base & 3.30 M & 0.620$_{\pm 0.263}$ & 0.771$_{\pm 0.194}$ & 0.776$_{\pm 0.195}$ & \textbf{0.793$_{\pm 0.182}$} \\
    NanoTabICL small-prior, small & 0.85 M & 0.461$_{\pm 0.136}$ & 0.719$_{\pm 0.192}$ & \textbf{0.745$_{\pm 0.181}$} & 0.742$_{\pm 0.192}$ \\
    NanoTabPFN big-prior, base & 3.72 M & 0.428$_{\pm 0.133}$ & 0.703$_{\pm 0.191}$ & 0.694$_{\pm 0.194}$ & \textbf{0.720$_{\pm 0.189}$} \\
    NanoTabPFN big-prior, small & 1.13 M & 0.436$_{\pm 0.130}$ & 0.676$_{\pm 0.185}$ & 0.694$_{\pm 0.195}$ & \textbf{0.736$_{\pm 0.191}$} \\
    NanoTabPFN small-prior, base & 3.72 M & 0.441$_{\pm 0.110}$ & 0.701$_{\pm 0.195}$ & 0.705$_{\pm 0.194}$ & \textbf{0.723$_{\pm 0.191}$} \\
    NanoTabPFN small-prior, small & 1.13 M & 0.448$_{\pm 0.115}$ & 0.671$_{\pm 0.186}$ & 0.700$_{\pm 0.194}$ & \textbf{0.748$_{\pm 0.184}$} \\
    \midrule
    \rowcolor{gray!12} \textit{mean over 8 nano ckpts} & --- & 0.489$_{\pm 0.187}$ & 0.712$_{\pm 0.193}$ & 0.726$_{\pm 0.193}$ & \textbf{0.746$_{\pm 0.189}$} \\
    \rowcolor{gray!12} \textit{$\Delta$ vs Vanilla FT} & --- & -0.223 & --- & +0.014 & \textbf{+0.033} \\
    \midrule
    \rowcolor{blue!8} TabICL v2 & 27.05 M & \textbf{0.891$_{\pm 0.132}$} & --- & --- & --- \\
    \rowcolor{blue!8} TabPFN v2.6 & 10.73 M & \textbf{0.888$_{\pm 0.137}$} & --- & --- & --- \\
    \bottomrule
    \end{tabular}
}
\end{table}

\begin{table}[H]
\centering\footnotesize
\caption{Mean balanced accuracy on the CUMIDA-42 gene-expression benchmark with the \emph{Random-100} feature arm (100 uniformly-sampled columns; 42 datasets, 3 seeds $\times$ 3 folds; mean $\pm$ std across all per-fold evaluations per cell). \textbf{Bold} = best in row among the FT methods (Zero-Shot excluded). v2 ceiling rows are populated when SLURM jobs land; currently shown as \emph{(in flight)}.}
\label{tab:cumida42_random100}
\resizebox{\textwidth}{!}{\begin{tabular}{@{}lrcccc@{}}
\toprule
Model & Params & Zero-Shot & Vanilla FT & LoRA FT & \modelname \\
\midrule
NanoTabICL big-prior, base & 3.30 M & 0.521$_{\pm 0.207}$ & 0.679$_{\pm 0.193}$ & \textbf{0.709$_{\pm 0.195}$} & 0.706$_{\pm 0.193}$ \\
NanoTabICL big-prior, small & 0.85 M & 0.500$_{\pm 0.181}$ & 0.663$_{\pm 0.197}$ & 0.677$_{\pm 0.195}$ & \textbf{0.687$_{\pm 0.199}$} \\
NanoTabICL small-prior, base & 3.30 M & 0.593$_{\pm 0.247}$ & 0.712$_{\pm 0.202}$ & \textbf{0.736$_{\pm 0.197}$} & 0.717$_{\pm 0.197}$ \\
NanoTabICL small-prior, small & 0.85 M & 0.450$_{\pm 0.126}$ & 0.665$_{\pm 0.196}$ & \textbf{0.680$_{\pm 0.193}$} & 0.667$_{\pm 0.199}$ \\
NanoTabPFN big-prior, base & 3.72 M & 0.434$_{\pm 0.129}$ & 0.631$_{\pm 0.191}$ & 0.636$_{\pm 0.189}$ & \textbf{0.669$_{\pm 0.192}$} \\
NanoTabPFN big-prior, small & 1.13 M & 0.435$_{\pm 0.127}$ & 0.618$_{\pm 0.186}$ & 0.634$_{\pm 0.184}$ & \textbf{0.681$_{\pm 0.202}$} \\
NanoTabPFN small-prior, base & 3.72 M & 0.442$_{\pm 0.111}$ & 0.640$_{\pm 0.194}$ & 0.635$_{\pm 0.191}$ & \textbf{0.663$_{\pm 0.197}$} \\
NanoTabPFN small-prior, small & 1.13 M & 0.448$_{\pm 0.117}$ & 0.630$_{\pm 0.181}$ & 0.636$_{\pm 0.184}$ & \textbf{0.680$_{\pm 0.201}$} \\
\midrule
\rowcolor{gray!12} \textit{mean over 8 nano ckpts} & --- & 0.478$_{\pm 0.171}$ & 0.655$_{\pm 0.194}$ & 0.668$_{\pm 0.194}$ & \textbf{0.684$_{\pm 0.198}$} \\
\rowcolor{gray!12} \textit{$\Delta$ vs Vanilla FT} & --- & -0.177 & --- & +0.013 & \textbf{+0.029} \\
\midrule
\rowcolor{blue!8} TabICL v2 & 27.05 M & \textbf{0.868$_{\pm 0.139}$} & 0.831$_{\pm 0.166}$ & 0.841$_{\pm 0.165}$ & 0.832$_{\pm 0.161}$ \\
\rowcolor{blue!8} TabPFN v2.6 & 10.73 M & \textbf{0.852$_{\pm 0.154}$} & 0.827$_{\pm 0.165}$ \textit{(28/42)} & 0.833$_{\pm 0.159}$ \textit{(27/42)} & 0.820$_{\pm 0.174}$ \textit{(26/42)} \\
\bottomrule
\end{tabular}}
\end{table}

\paragraph{What changes with feature-noise regime.} First, \modelname consistently improves over Vanilla FT, but the gain is not simply proportional to the amount of noise. The improvement is strongest in the moderately noisy setting, where the curated graph adjacency appears to provide the most useful signal beyond the raw features. Second, the v2 ICL ceilings are only mildly affected by feature noise. Even when moving from the cleanest regime to the fully random-feature regime, TabICL~v2 and TabPFN~v2.6 lose only a few BACC points. This suggests that frontier ICL backbones are highly robust to feature-level noise. The gap between these models and the tuned nano backbones widens slightly as noise increases, but remains substantial in every regime. For the two regimes in which v2 fine-tuning has been run, \texttt{top50\_rnd50} (\Cref{tab:cumida42_top50rnd50}) and \texttt{rnd100} (\Cref{tab:cumida42_random100}), all three fine-tuning variants---Vanilla FT, LoRA FT, and \modelname---fall below the \emph{Zero-Shot} zero-shot ICL baseline. This is consistent with the small-target-FT regression effect discussed in RQ4: adapting a frontier model to a small CUMIDA panel can disturb, rather than improve on, its broad synthetic-task prior. At v2 scale, KG-aware adaptation reduces neither the gap to the ICL ceiling nor the regression caused by fine-tuning.

\subsection{Comparison to matched density random graph.}
\begin{figure}[h]
\centering
\includegraphics[width=0.8\linewidth]{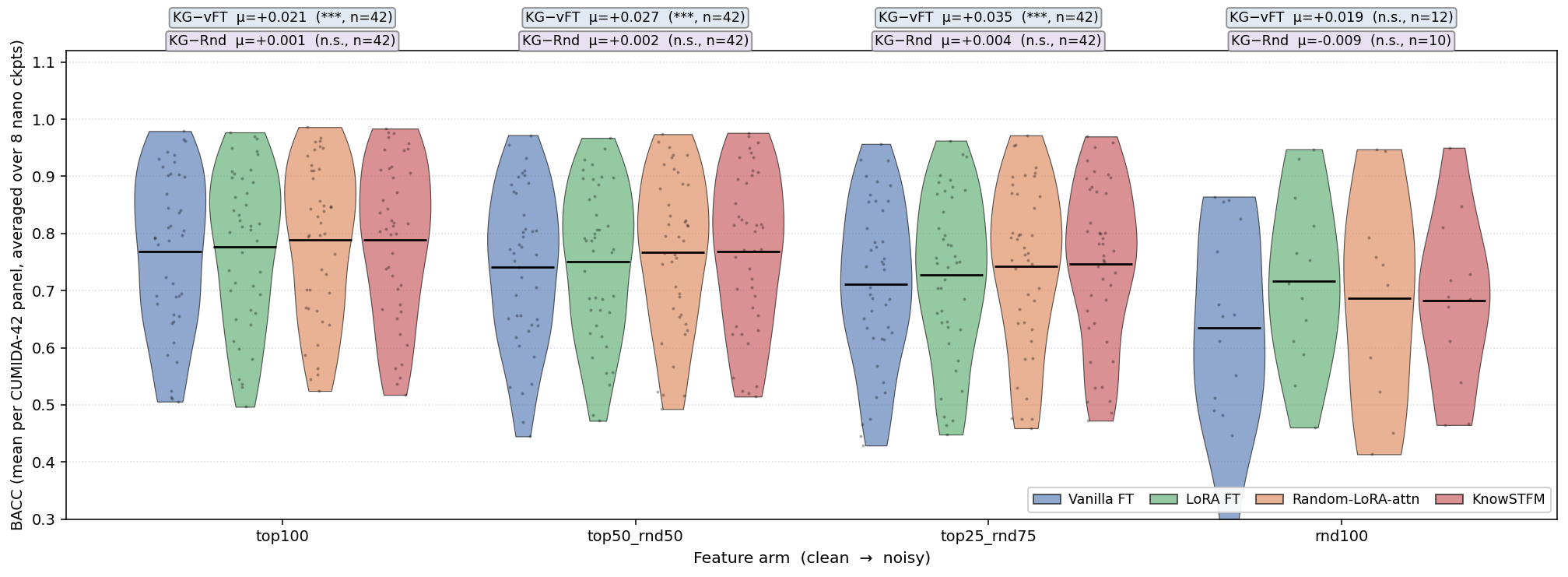}
\caption{Per-CUMIDA-42-panel BACC distributions (one violin per method per arm; jittered points are individual panels averaged over the 8 nano checkpoints). Each panel-level mean is shown as a horizontal tick. Annotation boxes report two-sided paired Wilcoxon signed-rank tests on the panel-level paired differences against Vanilla FT (\textsc{kg}$-$\textsc{vft}, blue) and against the matched-density random-graph control (\textsc{kg}$-$\textsc{rnd}, purple), with stars and paired sample size $n$. \modelname is significantly better than Vanilla FT on the three structured arms.}
\label{fig:cumida_arms_violin}
\end{figure}

To isolate the contribution of the PrimeKG signal from the inductive bias of injecting a sparse low-rank perturbation through the attention path, we add a matched-density random-graph control (\textbf{Rnd-LoRA}): an injection mask sampled uniformly at random with the same edge density as the PrimeKG-derived adjacency, kept fixed across folds and seeds, and trained under the identical LoRA schedule. Figure~\ref{fig:cumida_arms_violin} compares the per-panel BACC distributions of all four feature-attribution variants. \modelname improves significantly over Vanilla FT on every structured arm and earns the lowest mean rank on each variant, but at the panel level it is statistically tied with the matched-density random-graph control on all four arms. We read this as evidence that the dominant lever is the structural prior of routing fine-tuning through a sparse, low-rank, fixed attention mask, while the semantic content of the KG contributes a small directionally consistent residual that only becomes detectable in noisier feature regimes and at finer (checkpoint, dataset) granularity. We caution, however, that this control isolates edge-level structure conditional on an already domain-informed feature set: the candidate nodes in each panel are produced by per-fold Random Forest selection, which encodes substantial label-relevant biology before the graph is consulted. A complete decoupling of structure from semantics would additionally require a shuffled-mapping control that holds density and topology fixed while permuting the feature-to-node assignment, which we leave to future work.

\begin{table}[h]
\centering\footnotesize
\caption{Mean performance across the four CUMIDA-42 feature-attribution variants (rows). Each cell shows per-panel mean BACC ($\pm$ across-panel s.d.) above the autorank mean rank (lower is better; ranks computed per row over $n=42$ panels using \texttt{autorank}~\citep{Herbold2020}, non-parametric). Stars on the \modelname cell mark significance vs.\ Vanilla FT under a two-sided paired Wilcoxon signed-rank test: ${}^{***} p<10^{-3}$, ${}^{**} p<10^{-2}$, ${}^{*} p<0.05$, \textsc{ns} otherwise. \modelname achieves the best mean BACC and lowest rank on every variant, including the all-random stress arm.}
\label{tab:cumida_arms_pooled_summary}
\begin{tabular}{@{}l ccccc@{}}
\toprule
Variant & Zero-Shot & vFT & LoRA & Rnd-LoRA & \textbf{\modelname} \\
\midrule
\multirow{2}{*}{\texttt{top100} (clean)} & $0.505\,{\scriptstyle\pm 0.150}$ & $0.746\,{\scriptstyle\pm 0.136}$ & $0.777\,{\scriptstyle\pm 0.136}$ & $0.789\,{\scriptstyle\pm 0.136}$ & $\mathbf{0.789}\,{\scriptstyle\pm 0.138}$~${}^{***}$ \\
 & rank $4.92$ & rank $3.60$ & rank $2.67$ & rank $2.01$ & rank $\mathbf{1.81}$ \\
\midrule
\multirow{2}{*}{\texttt{top50\_rnd50} (headline)} & $0.501\,{\scriptstyle\pm 0.148}$ & $0.742\,{\scriptstyle\pm 0.137}$ & $0.752\,{\scriptstyle\pm 0.134}$ & $0.767\,{\scriptstyle\pm 0.137}$ & $\mathbf{0.768}\,{\scriptstyle\pm 0.140}$~${}^{***}$ \\
 & rank $4.90$ & rank $3.55$ & rank $2.90$ & rank $1.93$ & rank $\mathbf{1.71}$ \\
\midrule
\multirow{2}{*}{\texttt{top25\_rnd75} (noisy)} & $0.489\,{\scriptstyle\pm 0.141}$ & $0.711\,{\scriptstyle\pm 0.144}$ & $0.726\,{\scriptstyle\pm 0.142}$ & $0.742\,{\scriptstyle\pm 0.146}$ & $\mathbf{0.747}\,{\scriptstyle\pm 0.142}$~${}^{***}$ \\
 & rank $4.81$ & rank $3.73$ & rank $2.95$ & rank $1.95$ & rank $\mathbf{1.56}$ \\
\midrule
\multirow{2}{*}{\texttt{all-rnd} (noisy)} & $0.479\,{\scriptstyle\pm 0.135}$ & $0.633\,{\scriptstyle\pm 0.150}$ & $0.668\,{\scriptstyle\pm 0.146}$ & $0.681\,{\scriptstyle\pm 0.154}$ & $\mathbf{0.684}\,{\scriptstyle\pm 0.153}$~${}^{***}$ \\
 & rank $4.76$ & rank $3.62$ & rank $2.70$ & rank $2.04$ & rank $\mathbf{1.88}$ \\
\bottomrule
\end{tabular}
\end{table}

\newpage

%%% XDOMAIN13 

\section{Per-dataset cross-domain results}
\label{sec:appendix:xdomain13_per_dataset}

\subsection{Per-dataset HPO winners on \textsc{xdomain13}.}
\label{sec:xdomain13_kg_winners}

For each (checkpoint, dataset, fold, seed) we sweep $r \in \{8,16,32\}$, $\eta \in \{5\!\times\!10^{-4}, 1\!\times\!10^{-3}, 2\!\times\!10^{-3}\}$ and the eight three-block schedules $\boldsymbol{\sigma} \in \{\textsc{hard},\textsc{soft}\}^{3}$ (72 configs/split) and retain the val-best config per split; the test BACC reported below is the mean across the $n_\text{splits}{=}\,9$ resulting picks (3 seeds $\times$ 3 folds), matching the protocol used in Table~\ref{tab:xdomain13_adaptation_regimes}. Schedules are abbreviated H/S per block (e.g.\ \texttt{HSS} $= (\textsc{hard},\textsc{soft},\textsc{soft})$). The HPO landscape is flat: per cell the 9 splits typically elect $6$--$9$ distinct configs ($\#\text{cfg}$ column) and the modal config is rarely picked more than $2/9$ times. We therefore report the modal config and its frequency; the macro means $0.7298$ (\textsc{NanoTabICL}-base) and $0.7236$ (\textsc{NanoTabPFN}-base) are the per-fold val-best test averages and reproduce Table~\ref{tab:xdomain13_adaptation_regimes}.

\begin{table}[bt]
\centering
\caption{Per-dataset HPO winners on \textsc{xdomain13} for \texttt{kg\_lora\_attn\_FT} (paper protocol). \emph{Mode} is the modal val-best config across the 9 (seed, fold) splits with its frequency (out of 9); $\#\mathrm{cfg}$ counts how many distinct configs win at least one split. Test BACC is mean$\pm$sd over the same 9 splits, each evaluated at \emph{its own} val-best config. Macro averages match Table~\ref{tab:xdomain13_adaptation_regimes}.}
\label{tab:xdomain13_kg_winners}
\footnotesize
\setlength{\tabcolsep}{4pt}
\renewcommand{\arraystretch}{1.05}
\resizebox{!}{!}{%
\begin{tabular}{@{}llrllcc r@{}}
\toprule
Dataset & Backbone & $r$ & $\eta$ & $\boldsymbol{\sigma}$ & freq & test BACC & $\#\mathrm{cfg}$ \\
\midrule
\multirow{2}{*}{QSAR-biodeg}        & ICL  & 16 & $1\!\times\!10^{-3}$ & SHH & 1/9 & $0.833{\scriptstyle\pm0.026}$ & 9 \\
                                     & PFN  & 16 & $1\!\times\!10^{-3}$ & HSS & 2/9 & $0.813{\scriptstyle\pm0.025}$ & 7 \\
\midrule
\multirow{2}{*}{SPECT}              & ICL  & 32 & $5\!\times\!10^{-4}$ & SHH & 2/9 & $0.715{\scriptstyle\pm0.064}$ & 8 \\
                                     & PFN  & 16 & $2\!\times\!10^{-3}$ & HHS & 2/9 & $0.748{\scriptstyle\pm0.029}$ & 7 \\
\midrule
\multirow{2}{*}{automobile}         & ICL  & 16 & $1\!\times\!10^{-3}$ & SSH & 2/9 & $0.778{\scriptstyle\pm0.075}$ & 7 \\
                                     & PFN  &  8 & $1\!\times\!10^{-3}$ & SSS & 3/9 & $0.768{\scriptstyle\pm0.077}$ & 6 \\
\midrule
\multirow{2}{*}{cirrhosis}          & ICL  &  8 & $2\!\times\!10^{-3}$ & HSS & 1/9 & $0.491{\scriptstyle\pm0.022}$ & 9 \\
                                     & PFN  & 16 & $1\!\times\!10^{-3}$ & SHS & 2/9 & $0.531{\scriptstyle\pm0.051}$ & 8 \\
\midrule
\multirow{2}{*}{dermatology}        & ICL  & 16 & $5\!\times\!10^{-4}$ & SSS & 2/9 & $0.942{\scriptstyle\pm0.025}$ & 7 \\
                                     & PFN  & 32 & $1\!\times\!10^{-3}$ & SSS & 2/9 & $0.896{\scriptstyle\pm0.079}$ & 8 \\
\midrule
\multirow{2}{*}{diabetes\_pima}     & ICL  &  8 & $5\!\times\!10^{-4}$ & SHH & 2/9 & $0.699{\scriptstyle\pm0.024}$ & 7 \\
                                     & PFN  & 32 & $5\!\times\!10^{-4}$ & SHS & 1/9 & $0.719{\scriptstyle\pm0.032}$ & 9 \\
\midrule
\multirow{2}{*}{glass}              & ICL  & 16 & $2\!\times\!10^{-3}$ & HSH & 2/9 & $0.586{\scriptstyle\pm0.094}$ & 7 \\
                                     & PFN  & 32 & $1\!\times\!10^{-3}$ & SHS & 2/9 & $0.588{\scriptstyle\pm0.096}$ & 8 \\
\midrule
\multirow{2}{*}{glioma}             & ICL  & 32 & $2\!\times\!10^{-3}$ & SSH & 2/9 & $0.852{\scriptstyle\pm0.025}$ & 8 \\
                                     & PFN  &  8 & $1\!\times\!10^{-3}$ & SHS & 2/9 & $0.849{\scriptstyle\pm0.035}$ & 8 \\
\midrule
\multirow{2}{*}{heart-h}            & ICL  & 32 & $2\!\times\!10^{-3}$ & SSS & 2/9 & $0.776{\scriptstyle\pm0.047}$ & 8 \\
                                     & PFN  &  8 & $2\!\times\!10^{-3}$ & SHS & 2/9 & $0.782{\scriptstyle\pm0.022}$ & 7 \\
\midrule
\multirow{2}{*}{heart-statlog}      & ICL  & 16 & $2\!\times\!10^{-3}$ & SSH & 1/9 & $0.814{\scriptstyle\pm0.056}$ & 9 \\
                                     & PFN  & 16 & $1\!\times\!10^{-3}$ & SSS & 2/9 & $0.786{\scriptstyle\pm0.040}$ & 8 \\
\midrule
\multirow{2}{*}{hepatitis}          & ICL  & 32 & $1\!\times\!10^{-3}$ & HSS & 1/9 & $0.729{\scriptstyle\pm0.094}$ & 9 \\
                                     & PFN  &  8 & $2\!\times\!10^{-3}$ & HHS & 2/9 & $0.678{\scriptstyle\pm0.123}$ & 7 \\
\midrule
\multirow{2}{*}{steel-plates-fault} & ICL  & 32 & $2\!\times\!10^{-3}$ & HSH & 1/9 & $0.704{\scriptstyle\pm0.046}$ & 9 \\
                                     & PFN  & 16 & $2\!\times\!10^{-3}$ & SHS & 3/9 & $0.707{\scriptstyle\pm0.022}$ & 6 \\
\midrule
\multirow{2}{*}{stroke}             & ICL  &  8 & $1\!\times\!10^{-3}$ & HHH & 2/9 & $0.571{\scriptstyle\pm0.060}$ & 8 \\
                                     & PFN  &  8 & $1\!\times\!10^{-3}$ & HHS & 1/9 & $0.543{\scriptstyle\pm0.051}$ & 9 \\
\midrule
\multirow{2}{*}{\textbf{Mean (13 ds)}} & ICL & \multicolumn{5}{l}{} & $\mathbf{0.730}{\scriptstyle\,\pm0.125}$ \\
                                     & PFN & \multicolumn{5}{l}{} & $\mathbf{0.724}{\scriptstyle\,\pm0.113}$ \\
\bottomrule
\end{tabular}}
\end{table}

Each default cell is mean balanced accuracy across 9 seed-fold splits. The HPO columns report validation-selected Vanilla FT and KG-LoRA-attn sweeps where available.

\begin{table}[H]
\centering\scriptsize
\caption{BACC on \texttt{QSAR-biodeg} for big-prior base nano models and public TabICL/TabPFN.}
\label{tab:xdomain13_bigbase_QSAR_biodeg}
\begin{tabular}{@{}lcccccccc@{}}
\toprule
Model & Pretr. & vFT & LoRA & Rnd-LoRA & LLM-LoRA & KG-LoRA & vFT-HPO & KG-HPO \\
\midrule
NanoTabICL big/base & 0.499 & 0.817 & 0.826 & 0.831 & 0.804 & 0.831 & 0.808 & \textbf{0.833} \\
NanoTabPFN big/base & 0.500 & 0.818 & 0.818 & \textbf{0.828} & 0.542 & 0.806 & 0.812 & 0.812 \\
TabICL v2 & 0.848 & 0.821 & 0.833 & \textbf{0.841} & 0.834 & \textbf{0.841} & -- & -- \\
TabPFN v2.6 & 0.851 & 0.833 & \textbf{0.835} & 0.834 & 0.500 & 0.834 & -- & -- \\
\bottomrule
\end{tabular}
\end{table}

\begin{table}[H]
\centering\scriptsize
\caption{BACC on \texttt{SPECT} for big-prior base nano models and public TabICL/TabPFN.}
\label{tab:xdomain13_bigbase_SPECT}
\begin{tabular}{@{}lcccccccc@{}}
\toprule
Model & Pretr. & vFT & LoRA & Rnd-LoRA & LLM-LoRA & KG-LoRA & vFT-HPO & KG-HPO \\
\midrule
NanoTabICL big/base & 0.500 & \textbf{0.741} & 0.713 & 0.721 & 0.734 & 0.715 & 0.724 & 0.715 \\
NanoTabPFN big/base & 0.500 & 0.744 & 0.745 & 0.737 & 0.500 & 0.739 & 0.743 & \textbf{0.748} \\
TabICL v2 & 0.677 & 0.691 & 0.685 & 0.698 & \textbf{0.717} & 0.689 & -- & -- \\
TabPFN v2.6 & 0.678 & 0.718 & 0.715 & \textbf{0.734} & 0.500 & 0.710 & -- & -- \\
\bottomrule
\end{tabular}
\end{table}

\begin{table}[H]
\centering\scriptsize
\caption{BACC on \texttt{automobile} for big-prior base nano models and public TabICL/TabPFN.}
\label{tab:xdomain13_bigbase_automobile}
\begin{tabular}{@{}lcccccccc@{}}
\toprule
Model & Pretr. & vFT & LoRA & Rnd-LoRA & LLM-LoRA & KG-LoRA & vFT-HPO & KG-HPO \\
\midrule
NanoTabICL big/base & 0.500 & \textbf{0.852} & 0.811 & 0.779 & 0.500 & 0.788 & 0.802 & 0.778 \\
NanoTabPFN big/base & 0.500 & 0.716 & 0.680 & 0.630 & 0.500 & 0.591 & 0.730 & \textbf{0.768} \\
TabICL v2 & 0.881 & 0.819 & 0.856 & \textbf{0.875} & 0.665 & 0.862 & -- & -- \\
TabPFN v2.6 & 0.878 & 0.836 & 0.831 & 0.819 & 0.500 & \textbf{0.841} & -- & -- \\
\bottomrule
\end{tabular}
\end{table}

\begin{table}[H]
\centering\scriptsize
\caption{BACC on \texttt{cirrhosis} for big-prior base nano models and public TabICL/TabPFN.}
\label{tab:xdomain13_bigbase_cirrhosis}
\begin{tabular}{@{}lcccccccc@{}}
\toprule
Model & Pretr. & vFT & LoRA & Rnd-LoRA & LLM-LoRA & KG-LoRA & vFT-HPO & KG-HPO \\
\midrule
NanoTabICL big/base & 0.342 & 0.505 & 0.512 & 0.478 & 0.472 & 0.495 & \textbf{0.517} & 0.491 \\
NanoTabPFN big/base & 0.333 & 0.490 & 0.529 & 0.497 & 0.348 & 0.510 & 0.493 & \textbf{0.530} \\
TabICL v2 & 0.516 & 0.520 & \textbf{0.535} & 0.531 & 0.513 & 0.514 & -- & -- \\
TabPFN v2.6 & 0.510 & 0.510 & 0.502 & 0.503 & 0.333 & \textbf{0.529} & -- & -- \\
\bottomrule
\end{tabular}
\end{table}

\begin{table}[H]
\centering\scriptsize
\caption{BACC on \texttt{dermatology} for big-prior base nano models and public TabICL/TabPFN.}
\label{tab:xdomain13_bigbase_dermatology}
\begin{tabular}{@{}lcccccccc@{}}
\toprule
Model & Pretr. & vFT & LoRA & Rnd-LoRA & LLM-LoRA & KG-LoRA & vFT-HPO & KG-HPO \\
\midrule
NanoTabICL big/base & 0.167 & 0.922 & 0.936 & 0.940 & 0.915 & 0.931 & 0.917 & \textbf{0.942} \\
NanoTabPFN big/base & 0.167 & 0.900 & \textbf{0.914} & 0.911 & 0.167 & 0.899 & 0.910 & 0.896 \\
TabICL v2 & 0.976 & \textbf{0.974} & \textbf{0.973} & \textbf{0.974} & 0.972 & 0.968 & -- & -- \\
TabPFN v2.6 & 0.976 & 0.971 & \textbf{0.976} & 0.969 & 0.167 & 0.954 & -- & -- \\
\bottomrule
\end{tabular}
\end{table}

\begin{table}[H]
\centering\scriptsize
\caption{BACC on \texttt{diabetes\_pima} for big-prior base nano models and public TabICL/TabPFN.}
\label{tab:xdomain13_bigbase_diabetes_pima}
\begin{tabular}{@{}lcccccccc@{}}
\toprule
Model & Pretr. & vFT & LoRA & Rnd-LoRA & LLM-LoRA & KG-LoRA & vFT-HPO & KG-HPO \\
\midrule
NanoTabICL big/base & 0.502 & \textbf{0.718} & 0.714 & \textbf{0.718} & 0.715 & 0.713 & 0.700 & 0.699 \\
NanoTabPFN big/base & 0.500 & 0.699 & 0.713 & 0.712 & 0.663 & 0.718 & 0.704 & \textbf{0.719} \\
TabICL v2 & 0.715 & 0.710 & 0.710 & 0.701 & \textbf{0.727} & 0.718 & -- & -- \\
TabPFN v2.6 & 0.718 & 0.704 & \textbf{0.729} & 0.723 & 0.500 & 0.703 & -- & -- \\
\bottomrule
\end{tabular}
\end{table}

\begin{table}[H]
\centering\scriptsize
\caption{BACC on \texttt{glass} for big-prior base nano models and public TabICL/TabPFN.}
\label{tab:xdomain13_bigbase_glass}
\begin{tabular}{@{}lcccccccc@{}}
\toprule
Model & Pretr. & vFT & LoRA & Rnd-LoRA & LLM-LoRA & KG-LoRA & vFT-HPO & KG-HPO \\
\midrule
NanoTabICL big/base & 0.197 & 0.547 & \textbf{0.594} & 0.587 & 0.253 & 0.572 & 0.585 & 0.586 \\
NanoTabPFN big/base & 0.156 & 0.553 & 0.545 & 0.505 & 0.167 & 0.490 & 0.514 & \textbf{0.588} \\
TabICL v2 & 0.698 & 0.679 & 0.676 & \textbf{0.685} & 0.577 & 0.674 & -- & -- \\
TabPFN v2.6 & 0.713 & \textbf{0.692} & 0.674 & 0.668 & 0.167 & 0.658 & -- & -- \\
\bottomrule
\end{tabular}
\end{table}

\begin{table}[H]
\centering\scriptsize
\caption{BACC on \texttt{glioma} for big-prior base nano models and public TabICL/TabPFN.}
\label{tab:xdomain13_bigbase_glioma}
\begin{tabular}{@{}lcccccccc@{}}
\toprule
Model & Pretr. & vFT & LoRA & Rnd-LoRA & LLM-LoRA & KG-LoRA & vFT-HPO & KG-HPO \\
\midrule
NanoTabICL big/base & 0.500 & \textbf{0.854} & 0.843 & 0.853 & 0.636 & 0.847 & 0.850 & 0.852 \\
NanoTabPFN big/base & 0.500 & 0.849 & \textbf{0.865} & \textbf{0.865} & 0.500 & 0.859 & 0.856 & 0.849 \\
TabICL v2 & 0.877 & 0.856 & 0.866 & 0.867 & \textbf{0.870} & 0.855 & -- & -- \\
TabPFN v2.6 & 0.877 & \textbf{0.870} & 0.865 & 0.867 & 0.500 & \textbf{0.870} & -- & -- \\
\bottomrule
\end{tabular}
\end{table}

\begin{table}[H]
\centering\scriptsize
\caption{BACC on \texttt{heart-h} for big-prior base nano models and public TabICL/TabPFN.}
\label{tab:xdomain13_bigbase_heart_h}
\begin{tabular}{@{}lcccccccc@{}}
\toprule
Model & Pretr. & vFT & LoRA & Rnd-LoRA & LLM-LoRA & KG-LoRA & vFT-HPO & KG-HPO \\
\midrule
NanoTabICL big/base & 0.500 & 0.770 & 0.789 & 0.779 & 0.635 & \textbf{0.793} & 0.777 & 0.776 \\
NanoTabPFN big/base & 0.500 & 0.797 & 0.768 & 0.754 & 0.500 & 0.772 & \textbf{0.801} & 0.782 \\
TabICL v2 & 0.813 & 0.770 & 0.762 & 0.778 & 0.781 & \textbf{0.787} & -- & -- \\
TabPFN v2.6 & 0.800 & \textbf{0.777} & 0.771 & 0.764 & 0.500 & \textbf{0.777} & -- & -- \\
\bottomrule
\end{tabular}
\end{table}

\begin{table}[H]
\centering\scriptsize
\caption{BACC on \texttt{heart-statlog} for big-prior base nano models and public TabICL/TabPFN.}
\label{tab:xdomain13_bigbase_heart_statlog}
\begin{tabular}{@{}lcccccccc@{}}
\toprule
Model & Pretr. & vFT & LoRA & Rnd-LoRA & LLM-LoRA & KG-LoRA & vFT-HPO & KG-HPO \\
\midrule
NanoTabICL big/base & 0.513 & 0.775 & 0.799 & 0.799 & 0.807 & 0.809 & 0.800 & \textbf{0.814} \\
NanoTabPFN big/base & 0.500 & 0.776 & 0.791 & \textbf{0.802} & 0.611 & 0.786 & 0.778 & 0.786 \\
TabICL v2 & 0.857 & 0.810 & \textbf{0.832} & 0.828 & 0.803 & 0.813 & -- & -- \\
TabPFN v2.6 & 0.850 & 0.790 & 0.801 & 0.810 & 0.500 & \textbf{0.814} & -- & -- \\
\bottomrule
\end{tabular}
\end{table}

\begin{table}[H]
\centering\scriptsize
\caption{BACC on \texttt{hepatitis} for big-prior base nano models and public TabICL/TabPFN.}
\label{tab:xdomain13_bigbase_hepatitis}
\begin{tabular}{@{}lcccccccc@{}}
\toprule
Model & Pretr. & vFT & LoRA & Rnd-LoRA & LLM-LoRA & KG-LoRA & vFT-HPO & KG-HPO \\
\midrule
NanoTabICL big/base & 0.500 & \textbf{0.764} & 0.713 & 0.706 & 0.661 & 0.736 & 0.684 & 0.729 \\
NanoTabPFN big/base & 0.500 & 0.706 & 0.668 & 0.680 & 0.500 & \textbf{0.721} & 0.690 & 0.678 \\
TabICL v2 & 0.736 & \textbf{0.717} & 0.705 & 0.704 & 0.682 & 0.706 & -- & -- \\
TabPFN v2.6 & 0.741 & 0.721 & 0.704 & 0.715 & 0.500 & \textbf{0.732} & -- & -- \\
\bottomrule
\end{tabular}
\end{table}

\begin{table}[H]
\centering\scriptsize
\caption{BACC on \texttt{steel-plates-fault} for big-prior base nano models and public TabICL/TabPFN.}
\label{tab:xdomain13_bigbase_steel_plates_fault}
\begin{tabular}{@{}lcccccccc@{}}
\toprule
Model & Pretr. & vFT & LoRA & Rnd-LoRA & LLM-LoRA & KG-LoRA & vFT-HPO & KG-HPO \\
\midrule
NanoTabICL big/base & 0.156 & 0.703 & 0.691 & 0.705 & 0.373 & 0.705 & \textbf{0.708} & 0.704 \\
NanoTabPFN big/base & 0.143 & 0.659 & 0.694 & 0.667 & 0.155 & 0.661 & 0.664 & \textbf{0.707} \\
TabICL v2 & 0.821 & 0.781 & \textbf{0.797} & 0.786 & 0.745 & 0.786 & -- & -- \\
TabPFN v2.6 & 0.836 & \textbf{0.804} & 0.792 & 0.796 & 0.143 & 0.787 & -- & -- \\
\bottomrule
\end{tabular}
\end{table}

\begin{table}[H]
\centering\scriptsize
\caption{BACC on \texttt{stroke} for big-prior base nano models and public TabICL/TabPFN.}
\label{tab:xdomain13_bigbase_stroke}
\begin{tabular}{@{}lcccccccc@{}}
\toprule
Model & Pretr. & vFT & LoRA & Rnd-LoRA & LLM-LoRA & KG-LoRA & vFT-HPO & KG-HPO \\
\midrule
NanoTabICL big/base & 0.500 & 0.498 & 0.529 & 0.514 & 0.500 & 0.508 & 0.513 & \textbf{0.571} \\
NanoTabPFN big/base & 0.500 & 0.500 & 0.500 & 0.512 & 0.500 & 0.505 & 0.500 & \textbf{0.543} \\
TabICL v2 & 0.500 & 0.506 & 0.505 & 0.505 & 0.502 & \textbf{0.522} & -- & -- \\
TabPFN v2.6 & 0.500 & 0.521 & 0.519 & 0.534 & 0.500 & \textbf{0.544} & -- & -- \\
\bottomrule
\end{tabular}
\end{table}

\begin{table}[H]
\centering\scriptsize
\caption{Mean BACC across the 13 cross-domain datasets for big-prior base nano models and public TabICL/TabPFN.}
\label{tab:xdomain13_bigbase_summary}
\begin{tabular}{@{}lcccccccc@{}}
\toprule
Model & Pretr. & vFT & LoRA & Rnd-LoRA & LLM-LoRA & KG-LoRA & vFT-HPO & KG-HPO \\
\midrule
NanoTabICL big/base & 0.414 & 0.728 & 0.729 & 0.724 & 0.616 & 0.726 & 0.722 & \textbf{0.730} \\
NanoTabPFN big/base & 0.408 & 0.708 & 0.710 & 0.700 & 0.435 & 0.697 & 0.707 & \textbf{0.724} \\
TabICL v2 & 0.763 & 0.743 & 0.749 & \textbf{0.752} & 0.722 & 0.749 & -- & -- \\
TabPFN v2.6 & 0.764 & \textbf{0.750} & 0.747 & 0.749 & 0.408 & \textbf{0.750} & -- & -- \\
\bottomrule
\end{tabular}
\end{table}

\newpage

%%% APPENDIX WIKIDATA

\section{Per-dataset Wikidata mapping statistics across datasets}
\label{app:kg-stats}

For each of the 13 cross-domain datasets used in \Cref{tab:cross_domain_13ds}, we report the number of columns, mapping coverage (the fraction of columns to which the mapper of \Cref{sec:kg_mapping} assigned a non-empty QID), Wikidata edge counts by hop type (the union pool used in the headline adjacency $S^{(\rho)}$), the LLM-direct edge count, the resulting adjacency densities, and the Jaccard overlap between the two
feature--feature edge sets.

\begin{table}[h]
\centering
\footnotesize
\caption{
Mapping coverage and edge statistics for the 13 cross-domain datasets.
``cov.\,\%'' = fraction of columns to which the agentic mapper assigned a non-empty QID. ``direct/1-hop/ancestor'' = raw Wikidata edge counts under each relation policy; the headline adjacency takes their union. ``dens.\,\%'' = fraction of feature pairs connected under the Wikidata-union or LLM-direct adjacency. ``Jaccard\,\%'' = $100 \cdot |E_{\text{KG}} \cap E_{\text{LLM}}| / |E_{\text{KG}} \cup E_{\text{LLM}}|$ over feature-pair edges. Note: \texttt{SPECT} has $0\,\%$ KG density despite $1155$ raw ancestor edges because all 22 SPECT columns map to the same Wikidata entity (the \texttt{F1}--\texttt{F22} columns are abbreviated SPECT-image regions); \texttt{QSAR-biodeg} has $0$ KG edges because the agentic mapper abstained on 37 of 40 chemical-descriptor columns.
}
\begin{tabular}{@{}lrrrrrrrrr@{}}
\toprule
& & & \multicolumn{4}{c}{Wikidata KG (union)}
& \multicolumn{2}{c}{LLM-direct} & \\
\cmidrule(lr){4-7}
\cmidrule(lr){8-9}
Dataset
& $K$
& cov.\,\%
& direct
& 1-hop
& ancestor
& dens.\,\%
& $|E|$
& dens.\,\%
& Jaccard\,\% \\
\midrule
\texttt{automobile}         & 25 &  72 & 0 &   4 &  234 &  17.3 & 28 &  9.3 &  6.7 \\
\texttt{cirrhosis}          & 18 & 100 & 0 &   0 &  408 &  64.7 & 17 & 11.1 & 11.5 \\
\texttt{dermatology}        & 34 &  38 & 0 &   0 &  178 &   7.0 & 25 &  4.5 &  8.5 \\
\texttt{diabetes\_pima}     &  8 &  88 & 0 &   0 &   74 &  64.3 &  9 & 32.1 & 22.7 \\
\texttt{glass}              &  9 & 100 & 0 &  12 &  173 & 100.0 &  9 & 25.0 & 25.0 \\
\texttt{glioma}             & 23 & 100 & 0 &  13 &  980 &  92.1 & 24 &  9.5 & 10.3 \\
\texttt{heart-h}            & 13 &  92 & 0 &   2 &  193 &  56.4 & 14 & 17.9 & 16.0 \\
\texttt{heart-statlog}      & 13 &  85 & 0 &   2 &  199 &  55.1 & 13 & 16.7 & 16.7 \\
\texttt{hepatitis}          & 19 &  95 & 0 &   0 &  522 &  70.8 & 19 & 11.1 &  9.4 \\
\texttt{QSAR-biodeg}        & 40 &   8 & 0 &   0 &    0 &   0.0 & 31 &  4.0 &  0.0 \\
\texttt{SPECT}              & 22 & 100 & 0 &   0 & 1155 &   0.0 & 28 & 12.1 &  0.0 \\
\texttt{steel-plates-fault} & 27 &  78 & 0 &  50 &  393 &   8.8 & 25 &  7.1 &  5.7 \\
\texttt{stroke}             & 10 &  90 & 0 &   7 &   79 &  46.7 & 14 & 31.1 & 34.6 \\
\midrule
\textit{mean (13)}          & 20 &  79 & 0 &   7 &  354 &  44.9 & 20 & 14.7 & 12.8 \\
\bottomrule
\end{tabular}

\label{tab:kg_llm_stats}
\end{table}

\begin{figure}[h]
    \centering
    \includegraphics[width=\textwidth]{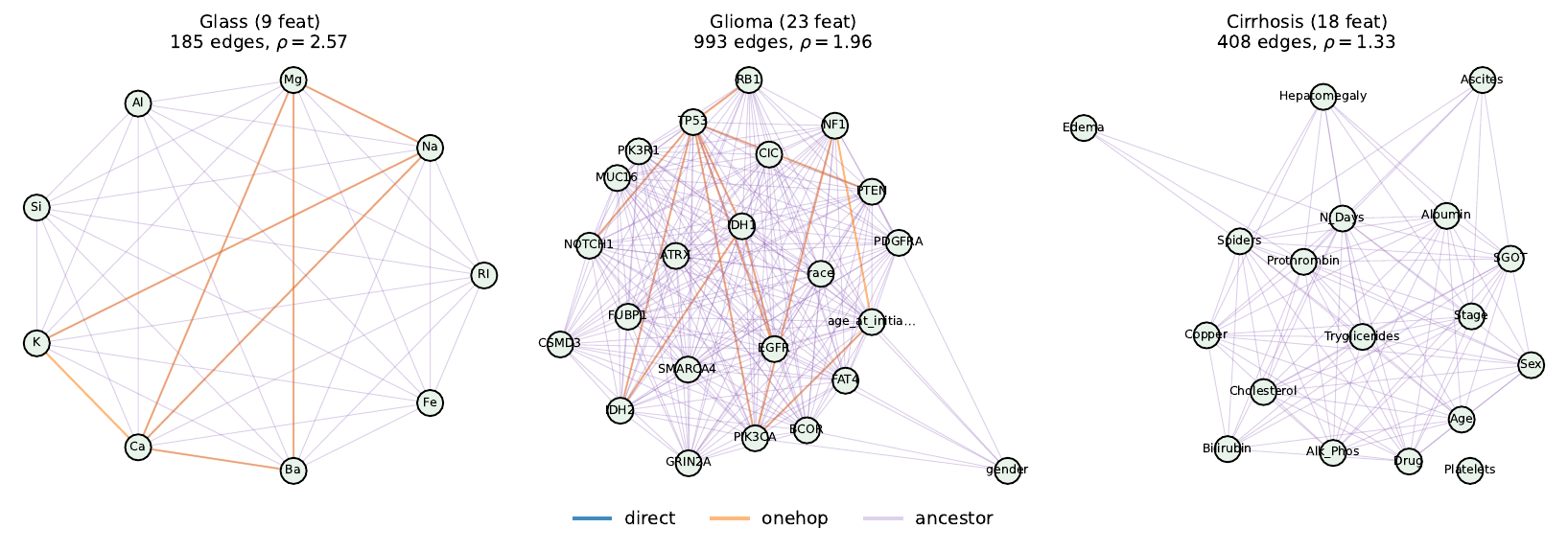}
    \caption{Wikidata adjacency graphs for the three densest \textsc{xdomain13} datasets (densest first by edges per ordered pair $\rho$): \emph{Glass} ($\rho{=}2.57$), \emph{Glioma} ($\rho{=}1.96$), and \emph{Cirrhosis} ($\rho{=}1.33$). Edges are coloured by Wikidata relation tier (\textcolor[HTML]{1F77B4}{direct}, \textcolor[HTML]{FF7F0E}
        {1-hop}, \textcolor[HTML]{9467BD}{ancestor}); the union over all tiers is the adjacency consumed by the \modelname injector. Nodes are dataset features placed by Kamad a--Kawai layout. The dominant tier is \emph{ancestor} (ontology parent shared via Wikidata superclass chains), explaining why $\rho > 1$: each pair carries multiple typed edges. }
    \label{fig:placeholder}
\end{figure}

\begin{figure}[t]                     
    \centering
    \includegraphics[width=\linewidth]{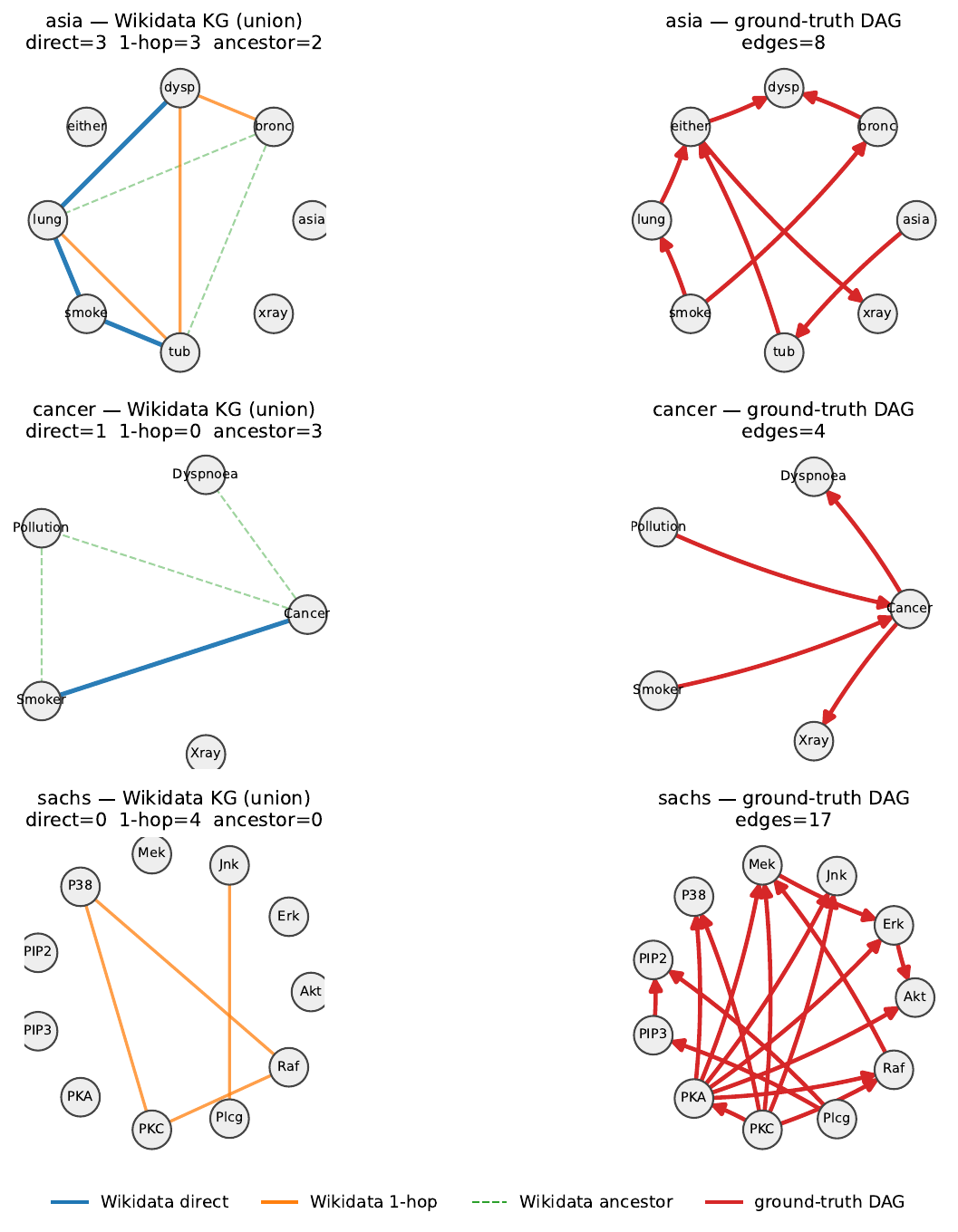}
    \caption{Wikidata-union KG vs.\ ground-truth DAG on \texttt{asia}, \texttt{cancer}, \texttt{sachs}. Edge colour = Wikidata hop tier (direct / 1-hop / ancestor); red 
  = DAG. Compact inset variant of the full figure.}                                                                                                                      
    \label{fig:struct_kg_vs_dag}
  \end{figure}

%%% ATTENTION VIZ
\section{Additional Attention Visualizations}
\label{app:attention_viz}

Here we provide layer-wise attention visualizations for the adapted TabPFN model.

\begin{figure}[htbp]
    \centering
    % First Image: Layer 00 (Hard Masking)
    \begin{subfigure}[b]{\textwidth}
        \centering
        \includegraphics[width=0.85\linewidth]{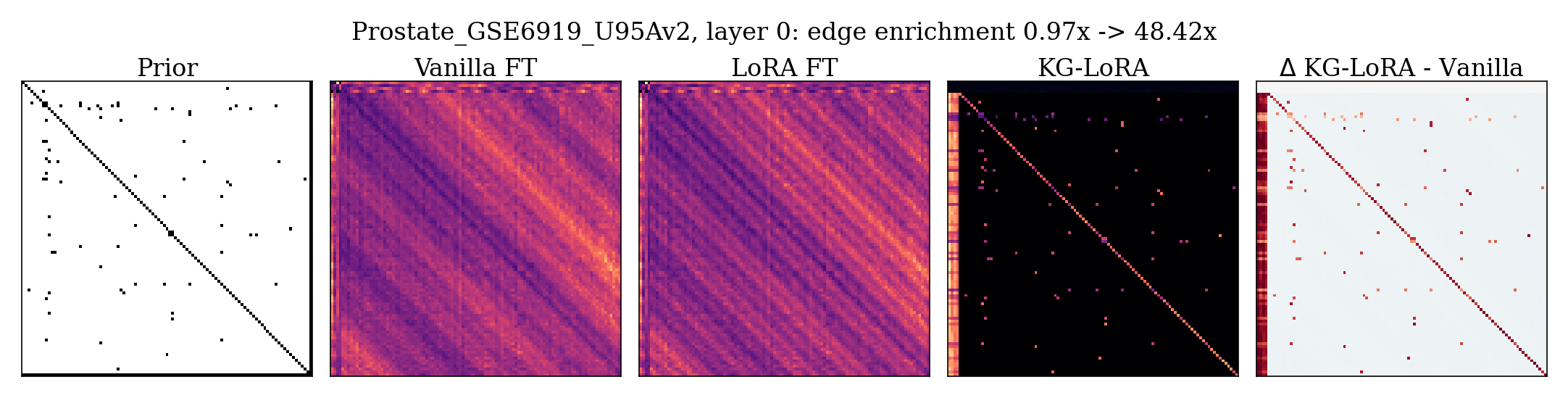}
        \caption{Layer 0. The initial hard masking strategy strictly constrains the attention weights, forcing the model to focus purely on the predefined structural priors resulting in a highly sparse pattern.}
        \label{fig:layer0}
    \end{subfigure}
    
    \vspace{0.5cm} 
    
    % Second Image: Layer 1 (Soft Masking)
    \begin{subfigure}[b]{\textwidth}
        \centering
        \includegraphics[width=0.85\linewidth]{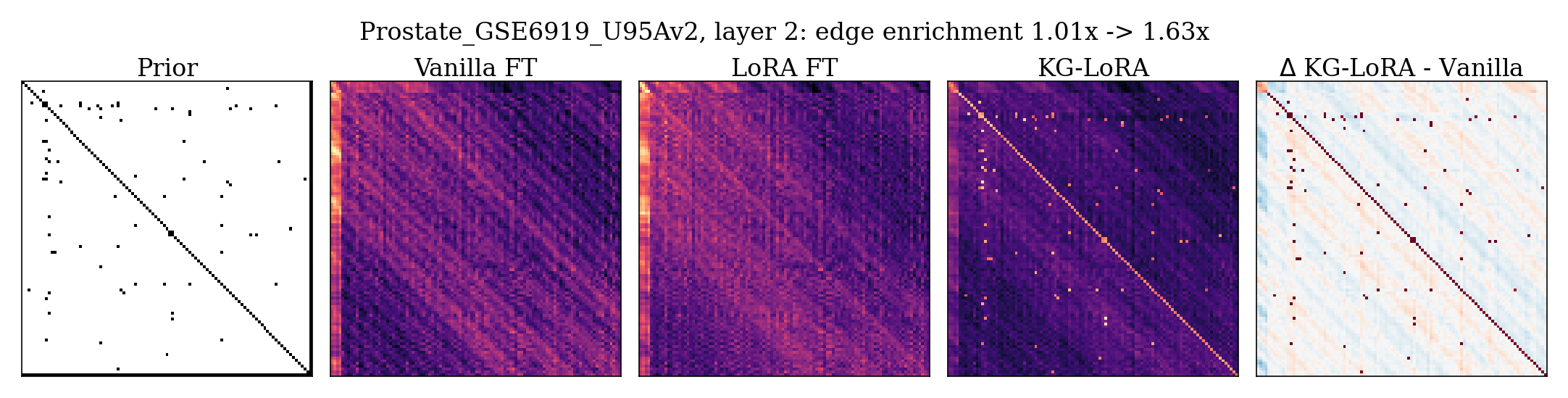}
        \caption{Layer 1. Transitioning to the first layer of soft masking, the model begins to distribute its attention more broadly, successfully blending the injected structural knowledge with learned semantic context.}
        \label{fig:layer1}
    \end{subfigure}
    
    \vspace{0.5cm}
    
    % Third Image: Layer 2 (Soft Masking)
    \begin{subfigure}[b]{\textwidth}
        \centering
        \includegraphics[width=0.85\linewidth]{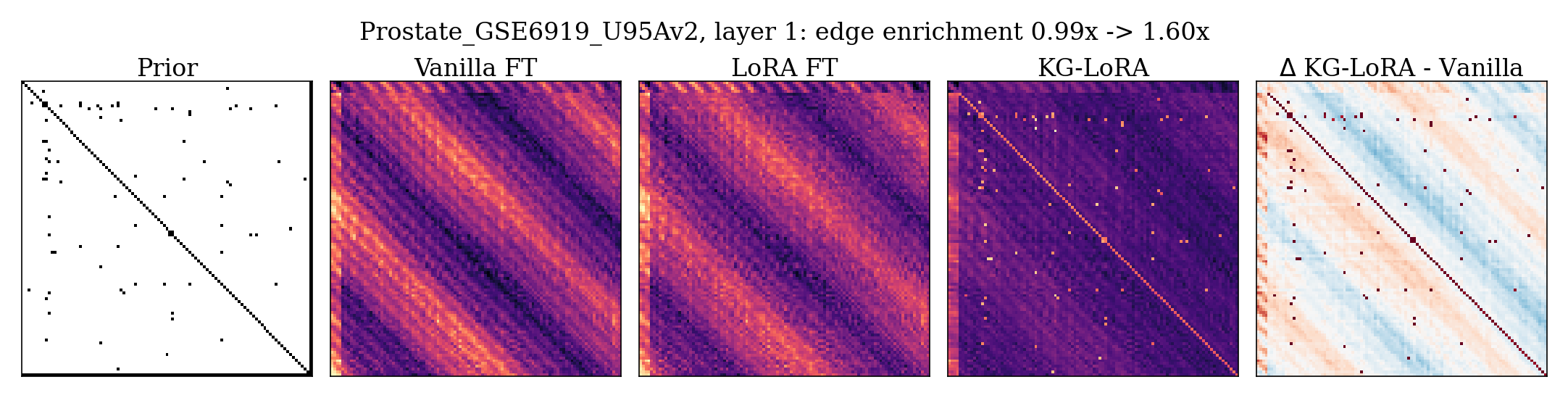}
        \caption{Layer 2. Continued soft masking in the deeper layers reveals a highly distributed attention pattern, indicating that the model is synthesizing higher-level, global relationships across the input.}
        \label{fig:layer2}
    \end{subfigure}
    
    \caption{Attention visualization of the TabPFN model on the Prostate-GSE6919 dataset with hard, soft, soft masking.}
    \label{fig:attn_appendix_stack}
\end{figure}

% \newpage
% \input{checklist.tex}

\end{document}